\pdfoutput=1

\documentclass[11pt]{article}

\usepackage[preprint]{acl}

\usepackage{times}
\usepackage{latexsym}

\usepackage[T1]{fontenc}

\usepackage[utf8]{inputenc}

\usepackage{microtype}

\usepackage{inconsolata}

\usepackage{graphicx}
\usepackage{todonotes}

\usepackage{colortbl}
\title{Mining Reasons For And Against Vaccination From Unstructured Data Using \textit{Nichesourcing} and AI Data Augmentation}

\author{
 \textbf{Damián Ariel Furman\textsuperscript{1,2}},
 \textbf{Juan Junqueras\textsuperscript{1}},
 \textbf{Z. Burçe Gümüşlü\textsuperscript{3}},
\\
 \textbf{Edgar Altszyler\textsuperscript{1,4}},
 \textbf{Joaquin Navajas\textsuperscript{5}},
 \textbf{Ophelia Deroy\textsuperscript{3}},
 \textbf{Justin Sulik\textsuperscript{3}},
\\
\\
 \textsuperscript{1} Universidad de Buenos Aires,
 \textsuperscript{2} Consejo Nacional de Investigaciones Científicas y Técnicas (CONICET),\\
 \textsuperscript{3} Ludwig-Maximilians-Universität München,
 \textsuperscript{4} Quantit,
 \textsuperscript{5} Universidad Torcuato Di Tella
\\
 \small{
   \textbf{Correspondence:} \href{damian.a.furman@gmail.com}{damian.a.furman@gmail.com}
 }
}

\begin{document}
\maketitle
\begin{abstract}
We present Reasons For and Against Vaccination (RFAV), a dataset for predicting reasons for and against vaccination, and scientific authorities used to justify them, annotated through \textit{nichesourcing} and augmented using GPT4 and GPT3.5-Turbo.
We show how it is possible to mine these reasons in non-structured text, under different task definitions, despite the high level of subjectivity involved and explore the impact of artificially augmented data using in-context learning with GPT4 and GPT3.5-Turbo.
We publish the \href{https://github.com/ArgMiningVaccination/RFAV-Dataset}{dataset} and the \href{https://huggingface.co/argmining-vaccines}{trained models}\footnote{https://huggingface.co/argmining-vaccines} along with the annotation manual used to train annotators and define the task\footnote{https://github.com/ArgMiningVaccination/RFAV-Dataset}.
\end{abstract}

\section{Introduction}

Over the last decades there had been an increase of anti-vaccine propaganda and parents deciding not to vaccinate their children, which have caused outbreaks of diseases that had been previously considered eliminated\cite{TAFURI20144860}. The massive development of Internet and communication technologies has provided a mean to facilitate information about vaccines and vaccination campaigns, but also a mean to spread misinformation\cite{KATA20101709}. In this scenario, the development of technologies for automatically recognising what is being said about vaccines can help to rapidly identify new misinformation campaigns and elaborate informed counter-narratives to mitigate the risk they pose.

In this work we present RFAV (Reasons For and Against Vaccination), a dataset with reasons for and against vaccination labeled through \textit{nichesourcing} and expanded using GPT4 and GPT3.5, on websites downloaded from different sources, in English and Spanish. We also trained different language models using this dataset to automatically identify reasons obtaining promising results. Since this task is highly subjective, we include an analysis of difficulties of the annotation process and assess the capabilities of generative LLMs for data augmentation in token classification tasks.

\section{Previous Work}

\citet{larson-hesitancy} defines Vaccine Hesitancy as "a state of indecision and uncertainty about vaccination before a decision is made (that) represents a time of vulnerability". \citet{Wilsone004206} presented a thorough study concluding that "there is a significant relationship between organization on social media and public doubts of vaccine safety".
In this sense, automatic tools "have potential to counter vaccine hesitancy"\cite{Larsonq69}, as they can help analyze massive online content. \citet{skeppstedt-etal} used topic models to manually code representative arguments about vaccines. \cite{qorib-etal} applied sentiment analysis for analyzing twitter user's stances about Covid-19 vaccines and reviewed other 14 studies that performed the same task. \citet{torsi-morante-2018-annotating} analyzed three annotation schemes for identifying argument components using a corpus of structured essays and news about vaccination and found that to achieve acceptable IAA they needed to use a simple scheme with only one component that was not strictly argumentative on itself. We follow a similar approach.

\section{Corpus creation}

\label{sec:datacollection}

To identify relevant web documents, we generated a list of \href{https://github.com/ArgMiningVaccination/RFAV-Dataset/blob/main/aux_files/keywords_SERAPI_en.json}{keywords} related to vaccination, including complementary/alternative medicine topics as these are associated with vaccine hesitance \cite{browne-etal}. We used SERAPI to conduct Google and Bing searches with those keywords, retrieving URLs for the top 150 hits per search. As this was a scattergun approach, we next sought to boost the proportion of relevant pages. We considered that any web domain reached by at least 10 unique keywords from our list was likely relevant, so we conducted additional SERPAPI searches focusing on those domains, retrieving up to 40 additional URLS per search per domain. Using the Trafilatura python package \cite{barbaresi-etal} we parsed the scraped HTML text for each URL, filtering out documents with fewer than 100 words. We used the TextDescriptives python package \cite{hansen-etal} to excise low-quality sections of text and the \href{https://microsoft.github.io/presidio/}{Presidio Analyzer} to sanitize personal identifying information. This yielded a total of 136934 documents in \href{https://github.com/ArgMiningVaccination/RFAV-Dataset/tree/main/raw_data/en}{English} and 94361 documents in \href{https://github.com/ArgMiningVaccination/RFAV-Dataset/tree/main/raw_data/es}{Spanish}. We further filtered these documents using a new list of \href{https://github.com/ArgMiningVaccination/RFAV-Dataset/blob/main/aux_files/Keywords.xlsx}{keywords} to preserve only those that were relevant to our purpose of annotation. After filtering, 94398 English documents (69\% of the corpus) and 66257 Spanish documents (70\% of the corpus) remained.


\subsection{Defining the task}
\label{sec:tasks}

All documents were labeled with Reasons for or against vaccination and with Scientific Authorities that might be used to support either a pro or an anti-vaccine stance within the document.

We define a Reason to be anything that can potentially be of interest to a person considering vaccination. They are not necessarily argumentative, though all arguments will be considered reasons. Each example may have zero-to-many reasons and each reason will be also labeled with a 'Stance' value using a Likert scale ranging from 1 to 5, defining the stance that the text has towards vaccination in a broad sense, according to the following descriptions:
\begin{enumerate}
    \vspace{-0.2cm}
    \item Strongly against vaccination
    \vspace{-0.2cm}
    \item Weakly against vaccination
    \vspace{-0.2cm}
    \item Ambiguous stance or undetermined
    \vspace{-0.2cm}
    \item Weakly supporting vaccination
    \vspace{-0.2cm}
    \item Strongly supporting vaccination
    \vspace{-0.2cm}
\end{enumerate}

Strong stances differ from Weak ones because they present themselves as conclusive and make their stance explicit. Weak stances, though relevant when considering vaccination, appear less conclusive and don't have an explicit posture.

We define a Scientific Authority to be any mention or invocation of scientists, publications, scientific, medical or governmental institutions used to provide credibility for potential reasons in the example (either for or against). The link between reasons and scientific authorities does not have to be explicit if it can be inferred that the authority is being cited to provide credibility.

More detailed descriptions of the categories defined and decisions about annotations with examples showing typical cases can be found in the  \href{https://github.com/ArgMiningVaccination/RFAV-Dataset/blob/main/aux_files/Annotation%20Manual%20for%20Mining%20Arguments%20For%20and%20Against%20Vaccination.docx}{Annotation Manual}.

In order to assess how the different categories of stances affect both human and machine performance, we define three tasks related to identification of Reasons: A - a per-word binary classification indicating if a word is part of a reason or not; B - a per-word classification using six categories: 0 for words not belonging to a reason and 1 to 5 to indicate stances; C - a per-word classification using four categories (0 to 3) consisting of a compressed version of the stances that doesn't account for the Weak vs Strong distinction. Considering also the task of predicting Scientific Authorities, this yields 4 different tasks.

\subsection{Data annotation}

We took a random sample of 1000 documents in English and 1000 examples in Spanish. We removed non-ascii characters and truncated the text to 4000 words, avoiding leaving unfinished sentences when possible.
Annotation was performed through \textit{nichesourcing} by six psychology and philosophy advanced college students divided in two teams for English and Spanish. Nichesourcing is "a specific form of outsourcing that harnesses the computational efforts from niche groups of experts rather than the ‘faceless crowd’"\cite{deboer-nichesourcing}.
Annotators were asked to carefully review the annotation manual and take a 2 hours course where vaccination-relevant concepts were explained and annotation criteria and examples were discussed.

Each of them then, labeled 400 examples: 100 were common to the three annotators in the same team while the other 300 were exclusive to each individual. This resulted in a total of 1000 examples labeled for each language, with 100 of those labeled three times used for calculating agreement.

Annotation was done using the brat annotation tool \cite{stenetorp-etal-2012-brat} in three stages. On the first and second stage all members of each team annotated 10 and 30 examples respectively from the common batch and did a pair-review discussing the cases where most disagreement arose. Criteria adopted on these stages was added to the annotation manual. On the third stage, all annotators from each team annotated the last 60 examples from the common batch and then the other 300 examples from their individual batch of examples.

\subsubsection{Agreement}
Agreement was calculated, for each language, using Cohen's $\kappa$\cite{cohen1960} between all possible pairs of annotators among the three members of the same language team. The reported score is the average of the three agreement values calculated for each combination of the three annotators.

Agreement was calculated in a per-word basis according to the four tasks defined in section \ref{sec:tasks}. For ``Reason'' and ``Scientific Authority'', agreement is calculated using a binary classification while for ``Stance'' and ``Compressed Stance'' is calculated for a multi-label classification with 6 and 4 classes respectively.

Table \ref{agreement_english} shows the agreement scores per component and per annotation round.

\begin{table}[]
\hspace{-0.5cm}
\begin{tabular}{lcccc}
\rowcolor[HTML]{C0C0C0} 
\textbf{ENGLISH}                                                   & R1                                           & R2                                           & R3  & All                 \\ \hline
\multicolumn{1}{|l|}{Reason}                                       & \multicolumn{1}{c|}{0.50}                         & \multicolumn{1}{c|}{0.49}                         & \multicolumn{1}{c|}{0.49} & \multicolumn{1}{c|}{0.49} \\ \hline
\multicolumn{1}{|l|}{Compressed stance}                            & \multicolumn{1}{c|}{0.40}                         & \multicolumn{1}{c|}{0.45}                         & \multicolumn{1}{c|}{0.43} & \multicolumn{1}{c|}{0.44} \\ \hline
\multicolumn{1}{|l|}{\cellcolor[HTML]{FFFFFF}Stance}               & \multicolumn{1}{c|}{\cellcolor[HTML]{FFFFFF}0.36} & \multicolumn{1}{c|}{\cellcolor[HTML]{FFFFFF}0.38} & \multicolumn{1}{c|}{0.36} & \multicolumn{1}{c|}{0.36} \\ \hline
\multicolumn{1}{|l|}{\cellcolor[HTML]{FFFFFF}Scientific Authority} & \multicolumn{1}{c|}{\cellcolor[HTML]{FFFFFF}0.41} & \multicolumn{1}{c|}{\cellcolor[HTML]{FFFFFF}0.20} & \multicolumn{1}{c|}{0.51} & \multicolumn{1}{c|}{0.45} \\ \hline
\hline
\hline
\rowcolor[HTML]{C0C0C0} 
\textbf{SPANISH}                                                   & R1                                             & R2                                           & R3 & All                  \\ \hline
\multicolumn{1}{|l|}{Reason}                                       & \multicolumn{1}{c|}{0.54}                           & \multicolumn{1}{c|}{0.50}                         & \multicolumn{1}{c|}{0.48} & \multicolumn{1}{c|}{0.49} \\ \hline
\multicolumn{1}{|l|}{Compressed stance}                            & \multicolumn{1}{c|}{0.54}                           & \multicolumn{1}{c|}{0.46}                         & \multicolumn{1}{c|}{0.47} & \multicolumn{1}{c|}{0.47} \\ \hline
\multicolumn{1}{|l|}{\cellcolor[HTML]{FFFFFF}Stance}               & \multicolumn{1}{c|}{\cellcolor[HTML]{FFFFFF}0.36}   & \multicolumn{1}{c|}{\cellcolor[HTML]{FFFFFF}0.39} & \multicolumn{1}{c|}{0.40} & \multicolumn{1}{c|}{0.39} \\ \hline
\multicolumn{1}{|l|}{\cellcolor[HTML]{FFFFFF}Scientific Authority} & \multicolumn{1}{c|}{\cellcolor[HTML]{FFFFFF}-0.002} & \multicolumn{1}{c|}{\cellcolor[HTML]{FFFFFF}0.16} & \multicolumn{1}{c|}{0.31} & \multicolumn{1}{c|}{0.25} \\ \hline
\end{tabular}
\caption{Cohen's Kappa agreement for English and Spanish. Table shows the average of the agreement between each possible pair of annotators from the three annotators for each language, divided between each round of annotation (1 to 3) and considering all three rounds}
\label{agreement_english}
\vspace{-0.5cm}
\end{table}

Based on Cohen's interpretation, Reason, Compressed Stances and Scientific Authority reach a moderate agreement, while Stance shows a fair agreement. For Spanish, Reason and Compressed Stances show a moderate agreement while Stance and Scientific Authority show a Fair agreement (being Stance, very close to moderate). The different values on each round of annotation show that even though the amount of examples was increased progressively, the level of agreement remains except for the case of Scientific Authority, where agreement improves with each round. This lead us to think that pair-reviews helped the annotators reach a better criteria.

Considering that annotation was performed on unstructured documents from different sources not necessarily vaccine-related, we consider this level of agreement to be satisfactory. Agreement is still on the same range of interpretation than \cite{poudyal-etal-2020-echr}, who achieved a Kappa agreement of 0.58 labeling arguments in a corpus of ECHR (European Court of Human Rights) decisions, considering they worked on more argumentatively structured examples. \cite{furman2023argumentative} labeled argumentative components as a binary classification obtaining an agreement score that ranges from 0.52 to 0.64 depending on the category. \citet{torsi-morante-2018-annotating} report 57\% annotator's match ratio on claim detection on debates about vaccination, using a metric ranging from 0 to 1 instead of Cohen's $\kappa$ that ranges from -1 to 1.

\subsubsection{Data Statistics}

\begin{figure}
    \centering
    \includegraphics[scale=0.35]{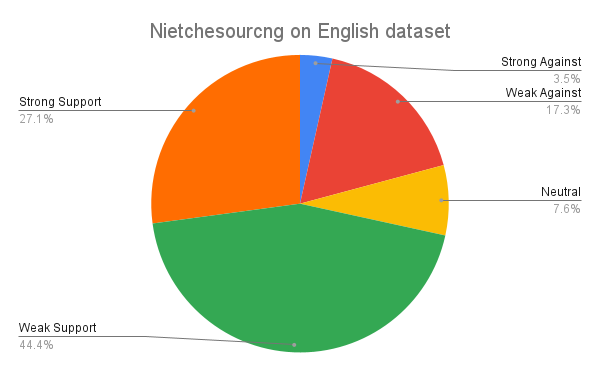}
    \includegraphics[scale=0.35]{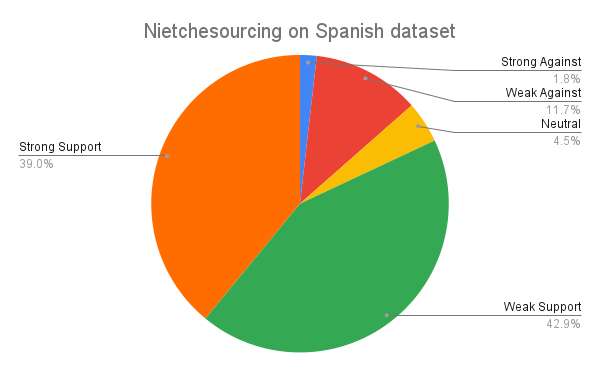}
    \caption{Distribution of labeled words per annotation class on English and Spanish expert annotated dataset}
    \label{fig:proportion_of_classes_en}
    \vspace{-0.5cm}
\end{figure}

Figure \ref{fig:proportion_of_classes_en} shows the distribution of words inside recognized reasons that were labeled for each class of Stance. Reasons supporting vaccination (either Strongly or Weakly) constitute 71.59\% of the total amount of Reasons labeled on the English dataset and 81,94\% on the Spanish dataset, while Reasons against vaccination are 20.76\% for English and 13.57\% for Spanish. Reasons Strongly Against vaccination are specially scarce in both Spanish and English.

Analyzing the data, we found that many of the documents that seemed to have been scraped from alternative medicine sources didn't mentioned vaccination and were filtered using the keywords as described in \ref{sec:datacollection}. We manually reviewed the 100 examples used for agreement calculation and found that most documents from these sources that mentioned vaccines avoided taking explicit stance. Some example of reasons against vaccination found are advertise possible secondary effects (sometimes selling treatment), not enforcing vaccination during Covid pandemic and, from a scientific perspective, also narrowing scope of vaccination campaigns.


\subsection{Data augmentation using GPT4 and GPT3.5Turbo}
\label{sec:dataaugmentation}


We used OpenAI's GPT4 to annotate 1000 new examples in each language and GPT3.5Turbo to annotate 2900 and 2400 new examples in English and Spanish respectively with Reasons and their Stances, spending U\$S600 on GPT4 and U\$S65 on GPT3.5-Turbo. 

We instructed the model to add [Begin:Reason:*Stance*] at the beginning of a reason and [End:Reason] at the end, where *Stance* stands for a value ranging from 1 to 5. The prompt also includes descriptions of the components to be annotated and instructions taken from the annotation manual, providing the model with similar information than human annotators. It also contains a three-shot learner with three labeled examples manually selected to contain reasons with diversity of stances.

While most examples annotated this way respected the proposed format, we found 11 cases in English and 6 cases in Spanish where the end token [End:Reason] was added before any start token. These cases were discarded and replaced with new ones.

Though the prompt instructed the model not to modify in any sense the original example, we noticed that GPT4 and GPT3.5Turbo sometimes introduced some minor changes like adding punctuation symbols at the beginning or at the end of the example, correction of orthography mistakes or syntactic errors or abruptly ending generation though not all words from the original example were processed.
We considered that these cases didn't constitute a significant change over the original example and found that the result of the annotation could be used without much problems for training models on all proposed tasks.

\subsubsection{Data Statistics}
\label{sec:gpt4_data_statistics}

\begin{table}[]
\begin{tabular}{l|ll||ll|}
\cline{2-5}
                                                                            & \multicolumn{2}{l||}{\cellcolor[HTML]{EFEFEF}Examples}                        & \multicolumn{2}{l|}{\cellcolor[HTML]{EFEFEF}Words labeled}                   \\ \cline{2-5} 
                                                                            & \multicolumn{1}{l|}{\cellcolor[HTML]{EFEFEF}EN} & \cellcolor[HTML]{EFEFEF}SP & \multicolumn{1}{l|}{\cellcolor[HTML]{EFEFEF}EN} & \cellcolor[HTML]{EFEFEF}SP \\ \hline
\multicolumn{1}{|l|}{\cellcolor[HTML]{EFEFEF}{\color[HTML]{333333} GPT4}}   & \multicolumn{1}{l|}{8.3\%}                      & 9.8\%                      & \multicolumn{1}{l|}{24\%}                       & 26\%                       \\ \hline
\multicolumn{1}{|l|}{\cellcolor[HTML]{EFEFEF}{\color[HTML]{333333} GPT3.5}} & \multicolumn{1}{l|}{12\%}                       & 14\%                       & \multicolumn{1}{l|}{24\%}                       & 24\%                       \\ \hline \hline
\multicolumn{1}{|l|}{\cellcolor[HTML]{EFEFEF}{\color[HTML]{333333} Human}}  & \multicolumn{1}{l|}{44.2\%}                     & 59.3\%                     & \multicolumn{1}{l|}{14\%}                       & 10\%                       \\ \hline
\end{tabular}
\caption{Percentage of examples with no Reason labeled (left) and percentage of words that formed part of a Reason (right) in English and Spanish for annotations using GPT4, GPT3.5 and \textit{nichesourcing}}
\label{tab:percentage-words-examples}
\end{table}

\begin{figure}
    \centering
    \includegraphics[scale=0.35]{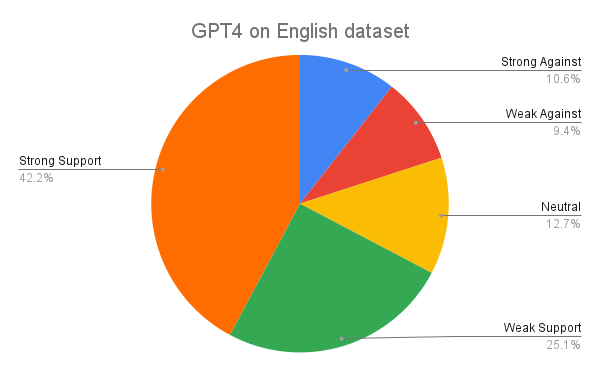}
    \includegraphics[scale=0.35]{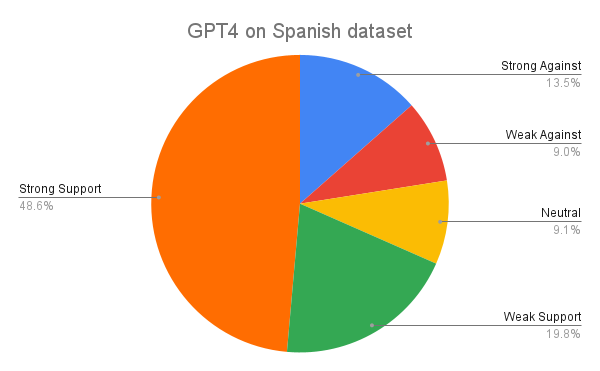}
    \includegraphics[scale=0.35]{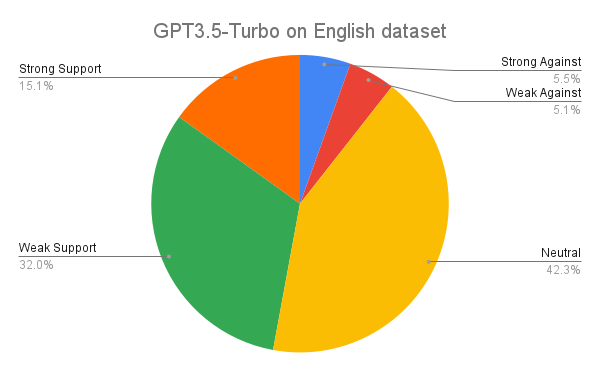}
    \includegraphics[scale=0.35]{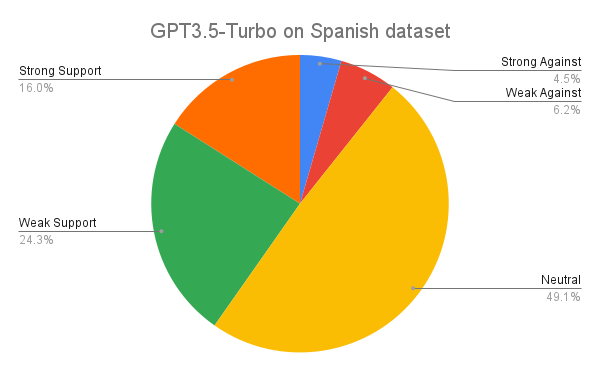}
    \caption{Distribution of labeled words per annotation class by GPT4 and GPT3.5-Turbno on English and Spanish datasets}
    \label{fig:proportion_of_classes_gpt4_en}
    \vspace{-0.5cm}
\end{figure}

Table \ref{tab:percentage-words-examples} shows the percentage of the examples that have no Reason labeled on them and also the percentage of words that are labeled as being part of a Reason, for English and Spanish and for corpus annotated through GPT4, GPT3.5-Turbo and \textit{nichesourcing} (humans). Though values are similar for GPT4 and GPT3.5-Turbo, it can be seen that human annotators labeled proportionally almost half the amount of reasons.
Figure \ref{fig:proportion_of_classes_gpt4_en} shows the distribution of words inside recognized reasons that were labeled for each class of Stance, for GPT4 and GPT3.5-Turbo and for English and Spanish respectively.
For GPT4, reasons supporting vaccination in the English dataset (either Strongly or Weakly) constitute 67.33\% of the total amount of Reasons while Reasons against vaccination are 19.96\%. In the Spanish dataset, reasons supporting vaccination labeled by GPT4 are 68.42\% while reasons against vaccination are 22.48\%.

For GPT3.5-Turbo, reasons supporting vaccination in the English dataset constitute 47.08\% of the total amount of Reasons while Reasons against vaccination are 10.60\%. In the Spanish dataset, reasons supporting vaccination labeled by GPT3.5-Turbo are 40.25\% while reasons against vaccination are 10.69\%.

In all cases, the proportion of Reasons labeled as "Strong Against" and specially the proportion of Reasons labeled as "Neutral" is much higher comparing to human annotators. In particular, for GPT3.5-Turbo, the Neutral class constitutes the majority class by a significant percentage (42.3\% for English and 49.1\% for Spanish), while the "Strong Support" that constitutes the majority class in all other datasets is greatly diminished in comparison.
We manually reviewed 20 examples that were not labeled with reasons by humans and found that GPT4 usually predicted sentences with a positive stance towards medical or scientific procedures in general as a "Support" reason and sentences with a positive stance towards Alternative Medicine related concepts as "Against" disregarding if they were referring to vaccination, while GPT3.5-Turbo usually labeled them as Neutral. Apart from that, we found many annotations that seemed to be reasonable but that differed with the criteria taken by the human annotator.


\section{Experiments}

Five pre-trained models (two in English, two in Spanish and one Multilingual) were fine-tuned using the English, Spanish and both portions of the corpus respectively, to automatize the tasks defined in \ref{sec:tasks}. Datasets were partitioned randomly in three parts for train, development and test, respecting a proportion of 80\%, 10\% and 10\% respectively. We explored three different learning rate values (2e-05, 1e-06 and 2e-06) and kept the model that had the best F1 score on the development partition.

We present a description of the models used:

\paragraph{RoBERTa}\cite{liu2019Roberta} is a transformer English language model based on BERT \cite{devlin-etal-2019-bert} that established a new state-of-the-art for 4 out of 9 GLUE tasks and matched state-of-the-art on other 2.

\paragraph{LongFormer} \cite{Beltagy2020Longformer} is a transformer based English language model specially designed for processing long documents. It is initialized using the weights of RoBERTa and then further pre-trained again on a corpus of 100K long documents to induce learning of long-range dependencies.

\paragraph{XLM Roberta} \cite{conneau2020unsupervised} is a transformer model based on Roberta architecture but trained with 2.5TB of data containing 100 different languages.

\paragraph{BETO} \cite{CaneteCFP2020} is a transformer based on BERT but trained from scratch from a big compilation of Spanish unannotated corpus from 15 different sources.

\paragraph{SpanBERTa}\cite{spanberta} is a model developed by SkimAI based on RoBERTa's architecture but trained from scratch with 18GB of Spanish data from a big corpus compiled from different sources.

\subsection{Training models with human annotated data}

\subsubsection{Evaluation}

For each model, we report F1, Precision and Recall scores over predictions on test dataset. For the two multi-label tasks of Stance recognition we also report F1 scores per category. Scores must be interpreted considering the subjective nature of the task so for the sake of comparison, we took the examples labeled by the three annotators (used to calculate agreement) and calculated F1 scores of all possible pairs considering one to be the ground truth and the other, a human predictor trying to mimic the other annotator (and vice-versa).
We use this score as an indicative measure of a machine predictor's performance compared to a human, per each category. We report the average of the six F1 scores calculated, the worst and the best scores, for English and Spanish.

\subsubsection{Results}
\label{sec:resultshuman}
Table \ref{tab:reason} shows the results obtained by all the classifiers trained for the task of Automatic Recognition of Reasons. For all models except BETO, performance was close or even slightly above Human annotators. Longformer performed best with an F1 score of 0.64. BETO performed more than 10 points below the other Spanish model, SpanBERTa.\\
Table \ref{tab:scientific_authorities} shows the results obtained by all the classifiers trained for the task of Automatic Recognition of cites of Scientific Authority. RoBERTa classifier obtained a 0.43 F1 score, above average human performance. Longformer classifier obtained a much lower score, demonstrating that long-range dependencies are not important for this task. For Spanish, Human performance is much lower, which corresponds to the lower agreement values shown in table \ref{agreement_english} so both models are above their performance. In this case, BETO obtained the higher score.
Table \ref{tab:stances} shows the results obtained by all the classifiers trained on the task of Automatic Recognition of Stances, predicting both if a word belongs to a reason and its Stance value. This is a difficult task because it involves classification using six labels with highly unbalanced distribution (see figure \ref{fig:proportion_of_classes_en}). Table \ref{tab:stance_classes} shows the F1 score per class for each model. English models achieve an acceptable performance for recognizing Support stances while they have no performance at all for Strong Against and Neutrals, both classes that were least frequents on the dataset. Only Longformer showed a better performance for Weak Against.\\
Table \ref{tab:compressed} shows the results obtained by all the classifiers trained in the task of Automatic Recognition of Compressed Stances. Results for all models rise between .11 and .14, a significant improvement, compared to the non-compressed version. This leads to interpret that a great amount of "mistakes" considered in the scoring of the models where because of difficulties at recognizing Strong vs Weak stances but not at recognizing Against vs Pro stances.\\
Unlike previous experiments, here the best model's performance (Longformer for English and SpanBERTa for Spanish) is .08 and .07 below human performance, respectively.\\
Table \ref{tab:compressed_classes} shows the F1 scores per class for this tasks. Spanish and Multilingual models still show no performance for Against class. Roberta, however, improved its performance significantly compared to both Against classes considered separately.\\
Pro class shows an improvement compared to both Pro classes from the the task of Stance recognition yielding a good performance, close to the binary task of recognizing reasons.\\
A manual examination of examples labeled by humans and by automatic models can be found on appendix section \ref{sec:manualexamexamples}.

\begin{table}[]
\vspace{-0.5cm}
\begin{tabular}{lrrr}
\rowcolor[HTML]{C0C0C0} 
Model                                                               & \multicolumn{1}{l}{\cellcolor[HTML]{C0C0C0}F1}       & \multicolumn{1}{l}{\cellcolor[HTML]{C0C0C0}Pr} & \multicolumn{1}{l}{\cellcolor[HTML]{C0C0C0}Rec} \\ \hline
\multicolumn{1}{|l|}{Roberta (EN)}                                  & \multicolumn{1}{r|}{0.56}                                  & \multicolumn{1}{r|}{0.69}                                   & \multicolumn{1}{r|}{0.47}                                \\ \hline
\multicolumn{1}{|l|}{Longformer (EN)}                                    & \multicolumn{1}{r|}{\textbf{0.64}}                         & \multicolumn{1}{r|}{0.58}                                   & \multicolumn{1}{r|}{0.72}                                \\ \hline
\rowcolor[HTML]{FFFFFF} 
\multicolumn{1}{|l|}{\cellcolor[HTML]{FFFFFF}XLM-Roberta (Multi)}      & \multicolumn{1}{r|}{\cellcolor[HTML]{FFFFFF}0.59}          & \multicolumn{1}{r|}{\cellcolor[HTML]{FFFFFF}0.66}           & \multicolumn{1}{r|}{\cellcolor[HTML]{FFFFFF}0.53}        \\ \hline
\rowcolor[HTML]{FFFFFF} 
\multicolumn{1}{|l|}{\cellcolor[HTML]{FFFFFF}SpanBERTa (SP)}              & \multicolumn{1}{r|}{\cellcolor[HTML]{FFFFFF}0.58}          & \multicolumn{1}{r|}{\cellcolor[HTML]{FFFFFF}0.47}           & \multicolumn{1}{r|}{\cellcolor[HTML]{FFFFFF}0.77}        \\ \hline
\multicolumn{1}{|l|}{\cellcolor[HTML]{FFFFFF}BETO (SP)}              & \multicolumn{1}{r|}{\cellcolor[HTML]{FFFFFF}0.47}          & \multicolumn{1}{r|}{\cellcolor[HTML]{FFFFFF}0.50}           & \multicolumn{1}{r|}{\cellcolor[HTML]{FFFFFF}0.44}        \\ \hline
\hline
\rowcolor[HTML]{EFEFEF} 
\multicolumn{1}{|l|}{Avg Human English} & \multicolumn{1}{r|}{\textbf{0.56}} & \multicolumn{1}{r|}{0.56}           & \multicolumn{1}{r|}{0.56}        \\ \hline
\rowcolor[HTML]{EFEFEF} 
\multicolumn{1}{|l|}{Best Human English}    & \multicolumn{1}{r|}{0.58}          & \multicolumn{1}{r|}{0.64}           & \multicolumn{1}{r|}{0.53}        \\ \hline
\rowcolor[HTML]{EFEFEF} 
\multicolumn{1}{|l|}{Worst Human English}   & \multicolumn{1}{r|}{0.52}          & \multicolumn{1}{r|}{0.5}            & \multicolumn{1}{r|}{0.54}        \\ \hline
\hline
\rowcolor[HTML]{EFEFEF}
\multicolumn{1}{|l|}{Avg Human Spanish} & \multicolumn{1}{r|}{\textbf{0.53}} & \multicolumn{1}{r|}{0.54}           & \multicolumn{1}{r|}{0.54}        \\ \hline
\rowcolor[HTML]{EFEFEF} 
\multicolumn{1}{|l|}{Best Human Spanish}    & \multicolumn{1}{r|}{0.57}          & \multicolumn{1}{r|}{0.71}           & \multicolumn{1}{r|}{0.47}        \\ \hline
\rowcolor[HTML]{EFEFEF} 
\multicolumn{1}{|l|}{Worst Human Spanish}   & \multicolumn{1}{r|}{0.49}          & \multicolumn{1}{r|}{0.46}           & \multicolumn{1}{r|}{0.54}        \\ \hline
\end{tabular}
\caption{F1, Precision and Recall scores of different models for the task of predicting reasons within an example}
\label{tab:reason}
\end{table}

\begin{table}[]
\vspace{-0.2cm}

\begin{tabular}{lrrr}
\rowcolor[HTML]{C0C0C0} 
Model                                                               & \multicolumn{1}{l}{\cellcolor[HTML]{C0C0C0}F1}       & \multicolumn{1}{l}{\cellcolor[HTML]{C0C0C0}Pr} & \multicolumn{1}{l}{\cellcolor[HTML]{C0C0C0}Rec} \\ \hline
\multicolumn{1}{|l|}{Roberta (EN)}                                       & \multicolumn{1}{r|}{\textbf{0.43}}                         & \multicolumn{1}{r|}{0.38}                                   & \multicolumn{1}{r|}{0.51}                                \\ \hline
\multicolumn{1}{|l|}{Longformer (EN)}                                    & \multicolumn{1}{r|}{0.29}                                  & \multicolumn{1}{r|}{0.68}                                   & \multicolumn{1}{r|}{0.19}                                \\ \hline
\rowcolor[HTML]{FFFFFF} 
\multicolumn{1}{|l|}{\cellcolor[HTML]{FFFFFF}XLM-Roberta (Multi)}           & \multicolumn{1}{r|}{\cellcolor[HTML]{FFFFFF}0.25}          & \multicolumn{1}{r|}{\cellcolor[HTML]{FFFFFF}0.49}           & \multicolumn{1}{r|}{\cellcolor[HTML]{FFFFFF}0.18}        \\ \hline
\rowcolor[HTML]{FFFFFF} 
\multicolumn{1}{|l|}{\cellcolor[HTML]{FFFFFF}SpanBERTa (SP)}              & \multicolumn{1}{r|}{\cellcolor[HTML]{FFFFFF}0.27}             & \multicolumn{1}{r|}{\cellcolor[HTML]{FFFFFF}0.46}              & \multicolumn{1}{r|}{\cellcolor[HTML]{FFFFFF}0.20}           \\ \hline
\multicolumn{1}{|l|}{\cellcolor[HTML]{FFFFFF}BETO (SP)}              & \multicolumn{1}{r|}{\cellcolor[HTML]{FFFFFF}0.36}          & \multicolumn{1}{r|}{\cellcolor[HTML]{FFFFFF}0.38}           & \multicolumn{1}{r|}{\cellcolor[HTML]{FFFFFF}0.33}        \\ \hline
\hline
\rowcolor[HTML]{EFEFEF} 
\multicolumn{1}{|l|}{Avg Human English} & \multicolumn{1}{r|}{\textbf{0.42}} & \multicolumn{1}{r|}{0.5}            & \multicolumn{1}{r|}{0.5}         \\ \hline
\rowcolor[HTML]{EFEFEF} 
\multicolumn{1}{|l|}{Best Human English}    & \multicolumn{1}{r|}{0.45}          & \multicolumn{1}{r|}{0.7}            & \multicolumn{1}{r|}{0.3}         \\ \hline
\rowcolor[HTML]{EFEFEF} 
\multicolumn{1}{|l|}{Worst Human English}   & \multicolumn{1}{r|}{0.38}          & \multicolumn{1}{r|}{0.25}           & \multicolumn{1}{r|}{0.83}        \\ \hline
\hline
\rowcolor[HTML]{EFEFEF} 
\multicolumn{1}{|l|}{Avg Human Spanish} & \multicolumn{1}{r|}{\textbf{0.25}} & \multicolumn{1}{r|}{0.28}           & \multicolumn{1}{r|}{0.28}        \\ \hline
\rowcolor[HTML]{EFEFEF} 
\multicolumn{1}{|l|}{Best Human Spanish}    & \multicolumn{1}{r|}{0.4}           & \multicolumn{1}{r|}{0.53}           & \multicolumn{1}{r|}{0.32}        \\ \hline
\rowcolor[HTML]{EFEFEF} 
\multicolumn{1}{|l|}{Worst Human Spanish}   & \multicolumn{1}{r|}{0.17}          & \multicolumn{1}{r|}{0.32}           & \multicolumn{1}{r|}{0.12}        \\ \hline
\end{tabular}
\caption{F1, Precision and Recall scores of different models for the task of predicting scientific authorities}
\label{tab:scientific_authorities}
\vspace{-0.6cm}
\end{table}

\begin{table}[]
\begin{tabular}{lrrr}
\rowcolor[HTML]{C0C0C0} 
Model                                                               & \multicolumn{1}{l}{\cellcolor[HTML]{C0C0C0}F1}       & \multicolumn{1}{l}{\cellcolor[HTML]{C0C0C0}Pr} & \multicolumn{1}{l}{\cellcolor[HTML]{C0C0C0}Rec} \\ \hline
\multicolumn{1}{|l|}{Roberta (EN)}                                  & \multicolumn{1}{r|}{0.28}                                  & \multicolumn{1}{r|}{0.33}                                   & \multicolumn{1}{r|}{0.28}                                \\ \hline
\multicolumn{1}{|l|}{Longformer (EN)}                                    & \multicolumn{1}{r|}{\textbf{0.31}}                         & \multicolumn{1}{r|}{0.35}                                   & \multicolumn{1}{r|}{0.30}                                \\ \hline
\rowcolor[HTML]{FFFFFF} 
\multicolumn{1}{|l|}{\cellcolor[HTML]{FFFFFF}XLM-Roberta (Multi)}      & \multicolumn{1}{r|}{\cellcolor[HTML]{FFFFFF}0.2}           & \multicolumn{1}{r|}{\cellcolor[HTML]{FFFFFF}0.26}           & \multicolumn{1}{r|}{\cellcolor[HTML]{FFFFFF}0.19}        \\ \hline
\rowcolor[HTML]{FFFFFF} 
\multicolumn{1}{|l|}{\cellcolor[HTML]{FFFFFF}SpanBERTa (SP)}              & \multicolumn{1}{r|}{\cellcolor[HTML]{FFFFFF}0.26}          & \multicolumn{1}{r|}{\cellcolor[HTML]{FFFFFF}0.26}           & \multicolumn{1}{r|}{\cellcolor[HTML]{FFFFFF}0.26}        \\ \hline
\multicolumn{1}{|l|}{\cellcolor[HTML]{FFFFFF}BETO (SP)}              & \multicolumn{1}{r|}{\cellcolor[HTML]{FFFFFF}0.24}          & \multicolumn{1}{r|}{\cellcolor[HTML]{FFFFFF}0.29}           & \multicolumn{1}{r|}{\cellcolor[HTML]{FFFFFF}0.23}        \\ \hline
\hline
\rowcolor[HTML]{EFEFEF} 
\multicolumn{1}{|l|}{Average Human English} & \multicolumn{1}{r|}{\textbf{0.36}} & \multicolumn{1}{r|}{0.38}           & \multicolumn{1}{r|}{0.38}        \\ \hline
\rowcolor[HTML]{EFEFEF} 
\multicolumn{1}{|l|}{Best Human English}    & \multicolumn{1}{r|}{0.54}          & \multicolumn{1}{r|}{0.53}           & \multicolumn{1}{r|}{0.54}        \\ \hline
\rowcolor[HTML]{EFEFEF} 
\multicolumn{1}{|l|}{Worst Human English}   & \multicolumn{1}{r|}{0.22}          & \multicolumn{1}{r|}{0.21}           & \multicolumn{1}{r|}{0.29}        \\ \hline
\hline
\rowcolor[HTML]{EFEFEF}
\rowcolor[HTML]{EFEFEF} 
\multicolumn{1}{|l|}{Average Human Spanish} & \multicolumn{1}{r|}{\textbf{0.31}} & \multicolumn{1}{r|}{0.33}           & \multicolumn{1}{r|}{0.33}        \\ \hline
\rowcolor[HTML]{EFEFEF} 
\multicolumn{1}{|l|}{Best Human Spanish}    & \multicolumn{1}{r|}{0.32}          & \multicolumn{1}{r|}{0.4}            & \multicolumn{1}{r|}{0.32}        \\ \hline
\rowcolor[HTML]{EFEFEF} 
\multicolumn{1}{|l|}{Worst Human Spanish}   & \multicolumn{1}{r|}{0.28}          & \multicolumn{1}{r|}{0.31}           & \multicolumn{1}{r|}{0.26}        \\ \hline
\end{tabular}
\caption{F1, Precision and Recall scores of different models for the task of predicting stances}
\label{tab:stances}
\end{table}

\begin{table}[]
\begin{tabular}{lrrr}
\rowcolor[HTML]{C0C0C0} 
Model                                                               & \multicolumn{1}{l}{\cellcolor[HTML]{C0C0C0}F1}       & \multicolumn{1}{l}{\cellcolor[HTML]{C0C0C0}Pr} & \multicolumn{1}{l}{\cellcolor[HTML]{C0C0C0}Rec} \\ \hline
\multicolumn{1}{|l|}{Roberta (EN)}                                  & \multicolumn{1}{r|}{0.43}                                  & \multicolumn{1}{r|}{0.48}                                   & \multicolumn{1}{r|}{0.41}                                \\ \hline
\multicolumn{1}{|l|}{Longformer (EN)}                                    & \multicolumn{1}{r|}{\textbf{0.43}}                         & \multicolumn{1}{r|}{0.48}                                   & \multicolumn{1}{r|}{0.4}                                 \\ \hline
\rowcolor[HTML]{FFFFFF} 
\multicolumn{1}{|l|}{\cellcolor[HTML]{FFFFFF}XLM-Roberta (Multi)}      & \multicolumn{1}{r|}{\cellcolor[HTML]{FFFFFF}0.36}          & \multicolumn{1}{r|}{\cellcolor[HTML]{FFFFFF}0.35}           & \multicolumn{1}{r|}{\cellcolor[HTML]{FFFFFF}0.38}        \\ \hline
\rowcolor[HTML]{FFFFFF} 
\multicolumn{1}{|l|}{\cellcolor[HTML]{FFFFFF}SpanBERTa (SP)}              & \multicolumn{1}{r|}{\cellcolor[HTML]{FFFFFF}0.36}          & \multicolumn{1}{r|}{\cellcolor[HTML]{FFFFFF}0.34}           & \multicolumn{1}{r|}{\cellcolor[HTML]{FFFFFF}0.39}        \\ \hline
\multicolumn{1}{|l|}{\cellcolor[HTML]{FFFFFF}BETO (SP)}              & \multicolumn{1}{r|}{\cellcolor[HTML]{FFFFFF}0.35}          & \multicolumn{1}{r|}{\cellcolor[HTML]{FFFFFF}0.35}           & \multicolumn{1}{r|}{\cellcolor[HTML]{FFFFFF}0.34}        \\ \hline
\hline
\rowcolor[HTML]{EFEFEF}
\multicolumn{1}{|l|}{Average Human English} & \multicolumn{1}{r|}{\textbf{0.51}} & \multicolumn{1}{r|}{0.51}           & \multicolumn{1}{r|}{0.51}        \\ \hline
\rowcolor[HTML]{EFEFEF} 
\multicolumn{1}{|l|}{Best Human English}    & \multicolumn{1}{r|}{0.54}          & \multicolumn{1}{r|}{0.53}           & \multicolumn{1}{r|}{0.56}        \\ \hline
\rowcolor[HTML]{EFEFEF} 
\multicolumn{1}{|l|}{Worst Human English}   & \multicolumn{1}{r|}{0.48}          & \multicolumn{1}{r|}{0.49}           & \multicolumn{1}{r|}{0.47}        \\ \hline
\hline
\rowcolor[HTML]{EFEFEF} 
\multicolumn{1}{|l|}{Average Human Spanish} & \multicolumn{1}{r|}{\textbf{0.43}} & \multicolumn{1}{r|}{0.44}           & \multicolumn{1}{r|}{0.44}        \\ \hline
\rowcolor[HTML]{EFEFEF} 
\multicolumn{1}{|l|}{Best Human Spanish}    & \multicolumn{1}{r|}{0.45}          & \multicolumn{1}{r|}{0.49}           & \multicolumn{1}{r|}{0.42}        \\ \hline
\rowcolor[HTML]{EFEFEF} 
\multicolumn{1}{|l|}{Worst Human Spanish}   & \multicolumn{1}{r|}{0.41}          & \multicolumn{1}{r|}{0.45}           & \multicolumn{1}{r|}{0.39}        \\ \hline 
\end{tabular}
\caption{F1, Precision and Recall scores of different models for the task of predicting a reduced set of stances (three instead of five)}
\label{tab:compressed}
\end{table}

\begin{table}[]\centering
\setlength\tabcolsep{4pt}
\begin{tabular}{l|l|l|l|l|l|}
\cline{2-6}
                                                    & \multicolumn{2}{|l|}{\cellcolor[HTML]{C0C0C0}Against}
                                                    & \multicolumn{1}{l|}{\cellcolor[HTML]{C0C0C0} Neu}
                                                    & \multicolumn{2}{l|}{\cellcolor[HTML]{C0C0C0}Support}\\ \hline
\rowcolor[HTML]{C0C0C0} 
\multicolumn{1}{|l|}{Model}                                                          & \multicolumn{1}{l|}{\cellcolor[HTML]{C0C0C0}Str} & \multicolumn{1}{l|}{\cellcolor[HTML]{C0C0C0}Wk} &                   & Wk              & Str            \\ \hline
\multicolumn{1}{|l|}{Roberta (EN)}                             & \multicolumn{1}{r|}{.0}                                   & \multicolumn{1}{r|}{.05}                                & \multicolumn{1}{l|}{.0} & \multicolumn{1}{l|}{.26} & \multicolumn{1}{l|}{.45} \\ \hline
\multicolumn{1}{|l|}{Longformer (EN)}                               & \multicolumn{1}{r|}{.0}                                   & \multicolumn{1}{r|}{.27}                                & \multicolumn{1}{l|}{.0} & \multicolumn{1}{l|}{.20} & \multicolumn{1}{l|}{.46} \\ \hline
\multicolumn{1}{|l|}{\cellcolor[HTML]{FFFFFF}XLM-Roberta (Multi)} & \multicolumn{1}{r|}{\cellcolor[HTML]{FFFFFF}.0}           & \multicolumn{1}{r|}{\cellcolor[HTML]{FFFFFF}.0}         & \multicolumn{1}{l|}{.0} & \multicolumn{1}{l|}{.14} & \multicolumn{1}{l|}{.14} \\ \hline
\multicolumn{1}{|l|}{\cellcolor[HTML]{FFFFFF}SpanBERTa (SP)}         & \multicolumn{1}{r|}{\cellcolor[HTML]{FFFFFF}.0}           & \multicolumn{1}{r|}{\cellcolor[HTML]{FFFFFF}.0}         & \multicolumn{1}{l|}{.0} & \multicolumn{1}{l|}{.31}  & \multicolumn{1}{l|}{.31}  \\ \hline
\multicolumn{1}{|l|}{\cellcolor[HTML]{FFFFFF}BETO (SP)}         & \multicolumn{1}{r|}{\cellcolor[HTML]{FFFFFF}.0}           & \multicolumn{1}{r|}{\cellcolor[HTML]{FFFFFF}.0}         & \multicolumn{1}{l|}{.0} & \multicolumn{1}{l|}{.21}  & \multicolumn{1}{l|}{.33}  \\ \hline
\end{tabular}
\caption{F1 Scores for the task of detecting stances per each class: Strong Against, Weak Against, Neutral, Weak Support and Strong Support}
\label{tab:stance_classes}
\end{table}

\begin{table}[]
\begin{tabular}{lrrl}
\rowcolor[HTML]{C0C0C0} 
Model                                                          & \multicolumn{1}{l}{\cellcolor[HTML]{C0C0C0}Against} & \multicolumn{1}{l}{\cellcolor[HTML]{C0C0C0}Neutral} & Pro                     \\ \hline
\multicolumn{1}{|l|}{Roberta (EN)}                             & \multicolumn{1}{r|}{.23}                           & \multicolumn{1}{r|}{.0}                            & \multicolumn{1}{l|}{.56} \\ \hline
\multicolumn{1}{|l|}{Longformer (EN)}                               & \multicolumn{1}{r|}{0.27}                           & \multicolumn{1}{r|}{.0}                            & \multicolumn{1}{l|}{.52} \\ \hline
\multicolumn{1}{|l|}{\cellcolor[HTML]{FFFFFF}XLM-Roberta (EN)} & \multicolumn{1}{r|}{\cellcolor[HTML]{FFFFFF}.0}    & \multicolumn{1}{r|}{\cellcolor[HTML]{FFFFFF}.0}    & \multicolumn{1}{l|}{.54} \\ \hline
\multicolumn{1}{|l|}{\cellcolor[HTML]{FFFFFF}SpanBERTa (SP)}         & \multicolumn{1}{r|}{\cellcolor[HTML]{FFFFFF}0.01}    & \multicolumn{1}{r|}{\cellcolor[HTML]{FFFFFF}.0}    & \multicolumn{1}{l|}{.50} \\ \hline
\multicolumn{1}{|l|}{\cellcolor[HTML]{FFFFFF}BETO (SP)}         & \multicolumn{1}{r|}{\cellcolor[HTML]{FFFFFF}0.00}    & \multicolumn{1}{r|}{\cellcolor[HTML]{FFFFFF}.0}    & \multicolumn{1}{l|}{.45} \\ \hline
\end{tabular}
\caption{F1 Scores per class for the task of detecting a compressed version of stances}
\label{tab:compressed_classes}
\vspace{-0.5cm}
\end{table}

\subsection{Training using augmented data}
\label{sec:gptresults}


Table \ref{tab:gpt_reason} shows the results of predictions of Reasons done by models trained by combining the dataset labeled through \textit{nichesourcing} with only GPT4 annotated dataset and with both GPT4 and GPT3.5-Turbo annotated datasets. Models were tested against the same test partition used for experiments in section \ref{sec:resultshuman}. It can be seen that by combining the Human annotated training partition with these datasets, the overall performance decreased. The more data we use for training the worse result we get. The models whose performance decreased the most are those who had a better performance when training only with the Human annotated corpus.
Tables \ref{tab:gpt_stances} and \ref{tab:gpt_compressed} show the results of training models with the same combination of datasets for predicting Stances and the Compressed Stances respectively. Again, we observe a decrease in model's performances but much smaller than when analysing Reasons.

\begin{table}[]
\vspace{-0.5cm}
\centering
\setlength\tabcolsep{4pt}
\begin{tabular}{l|lll||lll|}
\cline{2-7}
                                                    & \multicolumn{3}{l||}{\cellcolor[HTML]{C0C0C0}Hum + GPT4}                                                         & \multicolumn{3}{l|}{\cellcolor[HTML]{C0C0C0}All}\\ \hline
\rowcolor[HTML]{C0C0C0} 
\multicolumn{1}{|l|}{\cellcolor[HTML]{C0C0C0}Model} & \multicolumn{1}{l|}{\cellcolor[HTML]{C0C0C0}F1} & \multicolumn{1}{l|}{\cellcolor[HTML]{C0C0C0}Pr} & Rec & \multicolumn{1}{l|}{\cellcolor[HTML]{C0C0C0}F1} & \multicolumn{1}{l|}{\cellcolor[HTML]{C0C0C0}Pr} & Rec\\ \hline
\multicolumn{1}{|l|}{RoBERTa}                       & \multicolumn{1}{l|}{.45}                       & \multicolumn{1}{l|}{.70}                              & .33   & \multicolumn{1}{l|}{.31}                       & \multicolumn{1}{l|}{.71}                              & .20\\ \hline
\multicolumn{1}{|l|}{Longformer}                    & \multicolumn{1}{l|}{.39}                       & \multicolumn{1}{l|}{.78}                              & .26   & \multicolumn{1}{l|}{.19}                       & \multicolumn{1}{l|}{.83}                              & .11\\ \hline
\multicolumn{1}{|l|}{XLM-Roberta}                   & \multicolumn{1}{l|}{.52}                       & \multicolumn{1}{l|}{.54}                              & .50   & \multicolumn{1}{l|}{.10}                       & \multicolumn{1}{l|}{.78}                              & .05\\ \hline
\multicolumn{1}{|l|}{SpanBERTa}                     & \multicolumn{1}{l|}{.48}                       & \multicolumn{1}{l|}{.51}                              & .46   & \multicolumn{1}{l|}{.03}                       & \multicolumn{1}{l|}{.73}                              & .02\\ \hline
\multicolumn{1}{|l|}{BETO}                          & \multicolumn{1}{l|}{.43}                       & \multicolumn{1}{l|}{.61}                              & .33   & \multicolumn{1}{l|}{.20}                       & \multicolumn{1}{l|}{.83}                              & .11\\ \hline
\end{tabular}
\caption{Results of models trained with both Human + GPT4 and Human + GPT4 + GPT3.5 (All corpora) for predicting Reasons}
\label{tab:gpt_reason}
\end{table}

\begin{table}[]
\centering
\setlength\tabcolsep{4pt}
\begin{tabular}{l|lll||lll|}
\cline{2-7}
                                                    & \multicolumn{3}{l||}{\cellcolor[HTML]{C0C0C0}Hum + GPT4}                                                         & \multicolumn{3}{l|}{\cellcolor[HTML]{C0C0C0}All}                                                        \\ \hline
\rowcolor[HTML]{C0C0C0} 
\multicolumn{1}{|l|}{\cellcolor[HTML]{C0C0C0}Model} & \multicolumn{1}{l|}{\cellcolor[HTML]{C0C0C0}F1} & \multicolumn{1}{l|}{\cellcolor[HTML]{C0C0C0}Pr} & Rec & \multicolumn{1}{l|}{\cellcolor[HTML]{C0C0C0}F1} & \multicolumn{1}{l|}{\cellcolor[HTML]{C0C0C0}Pr} & Rec \\ \hline
\multicolumn{1}{|l|}{RoBERTa}                       & \multicolumn{1}{l|}{.27}                       & \multicolumn{1}{l|}{.28}                              & .29   & \multicolumn{1}{l|}{.21}                       & \multicolumn{1}{l|}{.23}                              & .22   \\ \hline
\multicolumn{1}{|l|}{Longformer}                    & \multicolumn{1}{l|}{.22}                       & \multicolumn{1}{l|}{.21}                              & .23   & \multicolumn{1}{l|}{.15}                       & \multicolumn{1}{l|}{.14}                              & .17   \\ \hline
\multicolumn{1}{|l|}{XLM-Roberta}                   & \multicolumn{1}{l|}{.27}                       & \multicolumn{1}{l|}{.26}                              & .28   & \multicolumn{1}{l|}{.18}                       & \multicolumn{1}{l|}{.24}                              & .18   \\ \hline
\multicolumn{1}{|l|}{SpanBERTa}                     & \multicolumn{1}{l|}{.23}                       & \multicolumn{1}{l|}{.22}                              & .24   & \multicolumn{1}{l|}{.21}                       & \multicolumn{1}{l|}{.26}                              & .22   \\ \hline
\multicolumn{1}{|l|}{BETO}                          & \multicolumn{1}{l|}{.20}                       & \multicolumn{1}{l|}{.34}                              & .19   & \multicolumn{1}{l|}{.22}                       & \multicolumn{1}{l|}{.48}                              & .20   \\ \hline
\end{tabular}
\caption{Results of models trained with both Human + GPT4 and Human + GPT4 + GPT3.5 (All corpora) for predicting Stances.}
\label{tab:gpt_stances}
\vspace{-0.5cm}
\end{table}

\begin{table}[]
\centering
\setlength\tabcolsep{4pt}
\begin{tabular}{l|lll||lll|}
\cline{2-7}
                                                    & \multicolumn{3}{l||}{\cellcolor[HTML]{C0C0C0}Hum + GPT4}                                                         & \multicolumn{3}{l|}{\cellcolor[HTML]{C0C0C0}All}                                                        \\ \hline
\rowcolor[HTML]{C0C0C0} 
\multicolumn{1}{|l|}{\cellcolor[HTML]{C0C0C0}Model} & \multicolumn{1}{l|}{\cellcolor[HTML]{C0C0C0}F1} & \multicolumn{1}{l|}{\cellcolor[HTML]{C0C0C0}Pr} & Rec & \multicolumn{1}{l|}{\cellcolor[HTML]{C0C0C0}F1} & \multicolumn{1}{l|}{\cellcolor[HTML]{C0C0C0}Pr} & Rec \\ \hline
\multicolumn{1}{|l|}{RoBERTa}                       & \multicolumn{1}{l|}{.42}                       & \multicolumn{1}{l|}{.49}                              & .39   & \multicolumn{1}{l|}{.32}                       & \multicolumn{1}{l|}{.59}                              & .32   \\ \hline
\multicolumn{1}{|l|}{Longformer}                    & \multicolumn{1}{l|}{.30}                       & \multicolumn{1}{l|}{.36}                              & .29   & \multicolumn{1}{l|}{.36}                       & \multicolumn{1}{l|}{.52}                              & .34   \\ \hline
\multicolumn{1}{|l|}{XLM-Roberta}                   & \multicolumn{1}{l|}{.38}                       & \multicolumn{1}{l|}{.43}                              & .36   & \multicolumn{1}{l|}{.23}                       & \multicolumn{1}{l|}{.40}                              & .25   \\ \hline
\multicolumn{1}{|l|}{SpanBERTa}                     & \multicolumn{1}{l|}{.38}                       & \multicolumn{1}{l|}{.40}                              & .37   & \multicolumn{1}{l|}{.31}                       & \multicolumn{1}{l|}{.37}                              & .29   \\ \hline
\multicolumn{1}{|l|}{BETO}                          & \multicolumn{1}{l|}{.35}                       & \multicolumn{1}{l|}{.52}                              & .33   & \multicolumn{1}{l|}{.35}                       & \multicolumn{1}{l|}{.58}                              & .32   \\ \hline
\end{tabular}
\caption{Results of models trained with both Human + GPT4 and Human + GPT4 + GPT3.5 (All corpora) for predicting Compressed Stances}
\label{tab:gpt_compressed}
\vspace{-1.5cm}
\end{table}

\begin{table}[]
\centering
\begin{tabular}{|l|l|l|l|}
\hline
\rowcolor[HTML]{C0C0C0} 
Model       & Hum & Hum+ GPT4 & All \\ \hline
RoBERTa     & 12.5\%     & 8.6\%        & 5.1\%                 \\ \hline
Longformer  & 22.6\%     & 6.1\%        & 2.4\%                 \\ \hline
XLM-Roberta & 14.6\%     & 12\%         & 1.2\%                 \\ \hline
SpanBERTa   & 21.5\%     & 3.9\%        & 0.3\%                 \\ \hline
BETO        & 11.5\%     & 7.2\%        & 1.8\%                 \\ \hline
\end{tabular}
\vspace{-0.2cm}
\caption{Percentage of words labeled by predictor as Reasons, for predictor trained with Human, Human + GPT4 and Human + GPT4 + GPT3.5Turbo annotated data. This is, the percentage of True positives + False positives over the whole dataset.}
\label{tab:gpt4_reasons_tp-fp}
\vspace{-0.5cm}
\end{table}

In order to gain insights for analyzing these results, we evaluated the performance of GPT4 and GPT3.5-Turbo against the test dataset using human annotations as gold standard. From 100 examples, 71 in English and 70 in Spanish were unchanged after adjusting the output with a postprocessing script that removes possible additions by GPT. The rest of the examples were discarded.

Tables \ref{tab:gpt4_performance_en} and \ref{tab:gpt4_performance_es} show F1, Precision and Recall scores for Automatic Recognition of Reasons, Stances and Compressed Stances for annotations done with GPT4 and compare those values to the ones obtained by the best model and human evaluation from section \ref{sec:resultshuman}.

These results seem to suggest that GPT models with fewshot learners don't perform as well as smaller open-source models finetuned with high quality data labeled by experts, or at least, they are not able to absorb the subjective criteria defined through the annotation process only by prompting and in-context learning.

Table \ref{tab:gpt4_reasons_tp-fp} shows the percentage of words that were labeled as being part of a Reason by each model. This is, the percentage of the annotated data that is either a True or a False positive. We can observe that the more data is used for training, the more conservative the model trained with that data becomes when predicting on the test dataset. This may seem contradictory on a first inspection given that the augmented data have almost twice as much positive labels than Human annotated examples (see section \ref{sec:gpt4_data_statistics}). Our hypothesis is that the combination of datasets with different annotation criteria affects negatively the models predictive capacity making them more conservative.

\begin{table}[]
\setlength\tabcolsep{4pt}
\begin{tabular}{|l|l|l|l||l|l|}
\hline
\rowcolor[HTML]{C0C0C0} 
Component       & F1   & Pr & Rec & Best F1 & Hum F1 \\ \hline
Reasons         & 0.43 & 0.44      & 0.44   & 0.64          & 0.56     \\ \hline
Stances         & 0.26 & 0.26      & 0.33   & 0.31          & 0.36     \\ \hline
Compressed         & 0.39 & 0.40      & 0.40   & 0.43          & 0.51     \\ \hline
\end{tabular}
\caption{Performance of GPT4 on the English test dataset for detecting reasons, stances and compressed stances, compared with the best model and human F1 scores for reference}
\label{tab:gpt4_performance_en}
\end{table}

\begin{table}[]
\setlength\tabcolsep{4pt}
\begin{tabular}{|l|l|l|l||l|l|}
\hline
\rowcolor[HTML]{C0C0C0} 
Component       & F1   & Pr & Rec & Best F1 & Hum F1 \\ \hline
Reasons         & 0.40 & 0.44      & 0.44   & 0.58          & 0.53     \\ \hline
Stances         & 0.23 & 0.27      & 0.27   & 0.26          & 0.31     \\ \hline
Compressed & 0.35 & 0.34      & 0.38   & 0.36          & 0.43     \\ \hline
\end{tabular}
\caption{Performance of GPT4 on the Spanish test dataset for detecting reasons, stances and compressed stances, compared with the best model and human F1 scores for reference}
\label{tab:gpt4_performance_es}
\vspace{-0.5cm}
\end{table}

\section{Conclusions}
In this work we present a protocol for annotating reasons with a stance towards vaccination and a dataset of 1000 examples in English and 1000 examples in Spanish annotated by six persons through \textit{Nichesourcing} and 3900 examples in English and 3400 examples in Spanish annotated using GPT4 and GPT3.5-Turbo with a fewshot learner and a short synthesis of the annotation manual on the prompt.
We release the dataset and the finetuned models for the free use of the scientific community.
Despite the highly subjective nature of the task, we achieved an acceptable IAA thanks to an iterative annotation process where annotation criteria and examples were discussed. Annotation manual registering this process is also released.
Experiments show that the annotation process can be reproduced automatically with satisfactory results considering the level of subjectivity of the correspondent task measured using Cohen's Kappa and the F1 scores of all combinations of annotators, with some room for improvement on the tasks of detecting Stances, particularly for the Against and Neutral classes.
When augmenting the human annotated corpus using annotations performed by GPT4 and GPT3.5-Turbo performance decreased, specially for the task of automatically identifying Reasons. Manual inspection of the augmented data revealed that the annotations made by GPT models were not senseless but rather followed a different criteria than human experts, tending to consider a wider range of subjects to be vaccine related (leading to annotate approximately 80\% more examples and twice the amount of words). We conclude that GPT models were not able to reproduce the annotation criteria of human annotators only by incorporating a reduced version of the annotation manual and three examples on the prompt.



\newpage
\section{Limitations}
In the following section we acknowledge some limitations found in our work.

Experiment results show that data imbalance of the annotated corpus directly affects the predictive capabilities of the models. Results from tables \ref{tab:stance_classes} and \ref{tab:stances} show that performance for the majority class ("Support") is much higher, while performance for the "Against" or "Neutral" classes is lower. This is related to the fact that these categories are scarce in the annotated dataset, as can be observed on figure \ref{fig:proportion_of_classes_en}. This data imbalance is a reflection of the proportion of online content supporting and attacking vaccination, being the first one much common than the second. Therefore, the only way to augment the sample of minority classes without artificially altering the distribution of classes is to label more examples, which is costly. This limits a possible use for the tool: to automatically recognize what is being said against vaccination in order to help elaborate adequate responses. More work needs to be done in order to improve model performance on minority classes.

Our strategy of data augmentation using generative models like GPT4 and GPT3.5-Turbo was based on providing them a summary of the annotation manual within the prompt and using in-context learning to make them learn the annotation criteria. However, annotation produced by these models followed a different criteria than human annotators, tending to consider a wider range of statements to be vaccine-related, therefore producing a different distribution of classes, which can be observed in figure \ref{fig:proportion_of_classes_gpt4_en}. While GPT4 tended to consider any statement that was science-related to be of the Pro class and any statement relative to alternative medicine to be of the Against class, GPT3.5 tended to annotate a lot of vaccine unrelated content as a Neutral Reason. We believe that this difference in annotation criteria made the models that were finetuned using this data combined with human annotations to become even more conservative in their labelling, specially over the minority classes, thus achieving lower results.


\section{Ethical Considerations}

Though this tool is intended to be used to fight misinformation campaigns causing vaccine hesitancy and possibly, outbreaks of preventable diseases, it could also be used as a tool to mine Reasons supporting vaccination in order to orient misinformation campaigns to target most commonly used reasons supporting vaccination. We acknowledge this possible misuse of our tool but we also reason that contrasting arguments, facts and information should help people to take more informed and rational decisions in the end.

Though one of our goals is to fight misinformation to help prevent outbreaks of preventable diseases, we also want to acknowledge that not all reasons against vaccination are necessarily misinformation. Example \ref{fig:immunosuppressed-patient} in appendix shows a reason against vaccination of immunosuppressed patients against COVID-19 based on lack of testing and the possibility to wait given that there was little cases in that country at that time. We found a significant amount of examples like this one, where reasons for not getting vaccinated were presented not against vaccination in general, but against a particular vaccine or vaccination campaign and they were presented with a scientific base. The dataset along with the trained models presented in this work are meant to help to automatically identify what is being said about vaccines and vaccination. It must be used with caution and critical thinking.

\newpage
\bibliography{custom}
\newpage
\appendix

\section{Prompt used for data augmentation}
Figure \ref{fig:template_prompt} shows the template used for generating the prompts. While this was the same for both languages, examples used for in-context learning were selected to match the same language as the example being annotated.

\begin{figure}
    \flushleft
    \small
    \texttt{Extract reasons either supporting or opposing vaccination, and link them to the corresponding stance values.\\
	• A Reason potentially or hypothetically appeals to someone considering vaccination.\\
	• They must be something relevant to someone hypothetically considering getting or not getting vaccinated.\\
	• Examples can have zero or many reasons.  \\
	• Reasons have a number indicating their stance towards vaccination.\\
	• The token [Reason:begin:1] indicates a reason that is strongly against vaccination.\\
	• The token [Reason:begin:2] indicates a reason that is weakly against vaccination, this means, that it highlights negative aspects associated with vaccination without explicitly taking a stance against it.\\
	• The token [Reason:begin:3] indicates a reason that have a neutral stance towards vaccination or which stance can not be inferred.\\
	• The token [Reason:begin:4] indicates a reason weakly supporting vaccination. This means that it provides positive aspects of vaccination (like "they are free" or "they are accessible") without explicitly taking a stance.\\
	• The token [Reason:begin:5] indicates a reason strongly supporting vaccination. It associates vaccines explicitly with good qualities and positive concepts.\\
	• Do not remove the links from the original non annotated text, keep them in plain text, respecting the original format.\\
	• An identified reason should be marked with the special tokens "[Reason:begin:stanceValue]" before the first word of the reason and "[Reason:end]" at the end of the reason.\\
	• stanceValue is an integer ranging from 1 to 5.
	• The output should contain exactly the same text as the input only adding the special tokens when appropriate.
}\\
    \vspace{-0.2cm}
    \caption{\small Template used for generating prompts for annotation using GPT4 and GPT3.5. The final version of the prompt included three non-annotated examples linked to their correspondent annotations}
    \label{fig:template_prompt}
\end{figure}

\section{Example Appendix}
\label{sec:appendix}

\subsection{Manual examination of predicted examples}
\label{sec:manualexamexamples}

\begin{figure*}
\centering
        \includegraphics[scale=0.25]{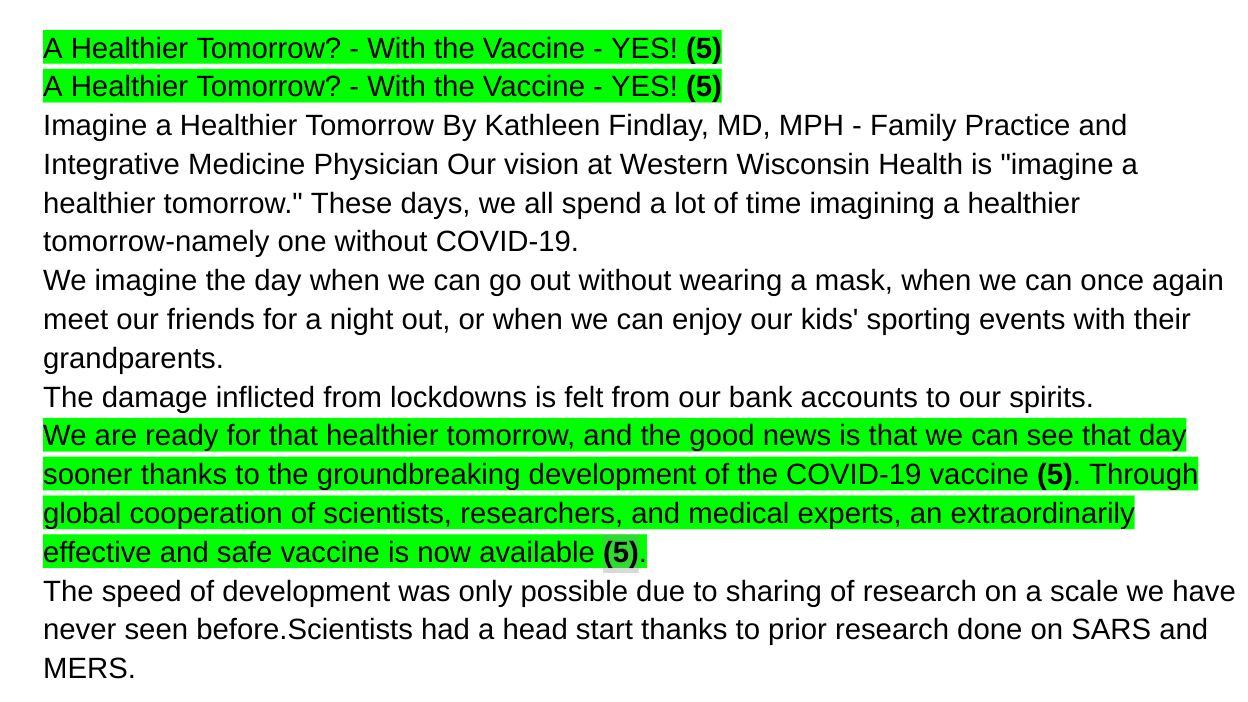}
        \caption{Example 105 from test dataset labeled through nichesourcing with Reasons, Stances and Scientific Authorities}
    \label{fig:example_A_human}
\end{figure*}

\begin{figure*}
\centering
        \includegraphics[scale=0.3]{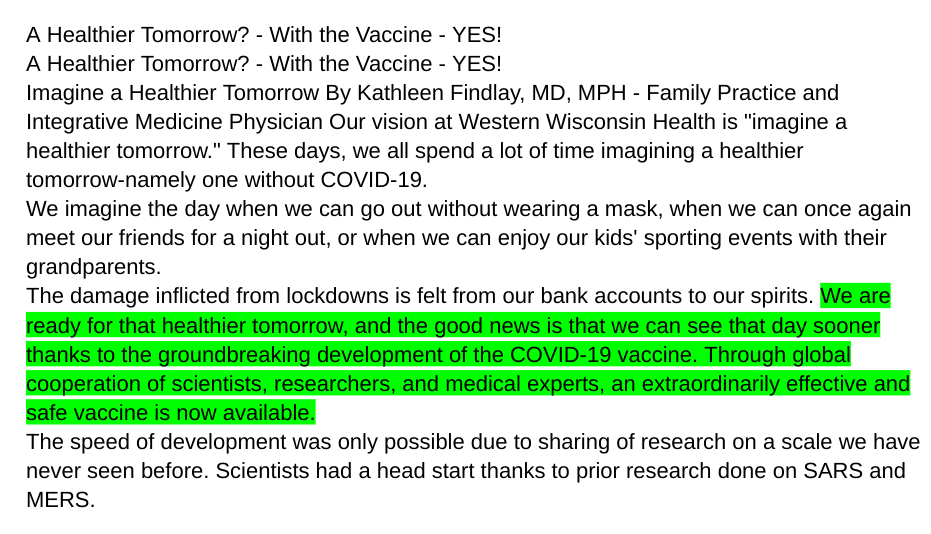}
        \caption{Example 105 from test dataset labeled by our finetuned Longformer only with Reasons}
    \label{fig:example_A_reason}
\end{figure*}

\begin{figure*}
\centering
        \includegraphics[scale=0.25]{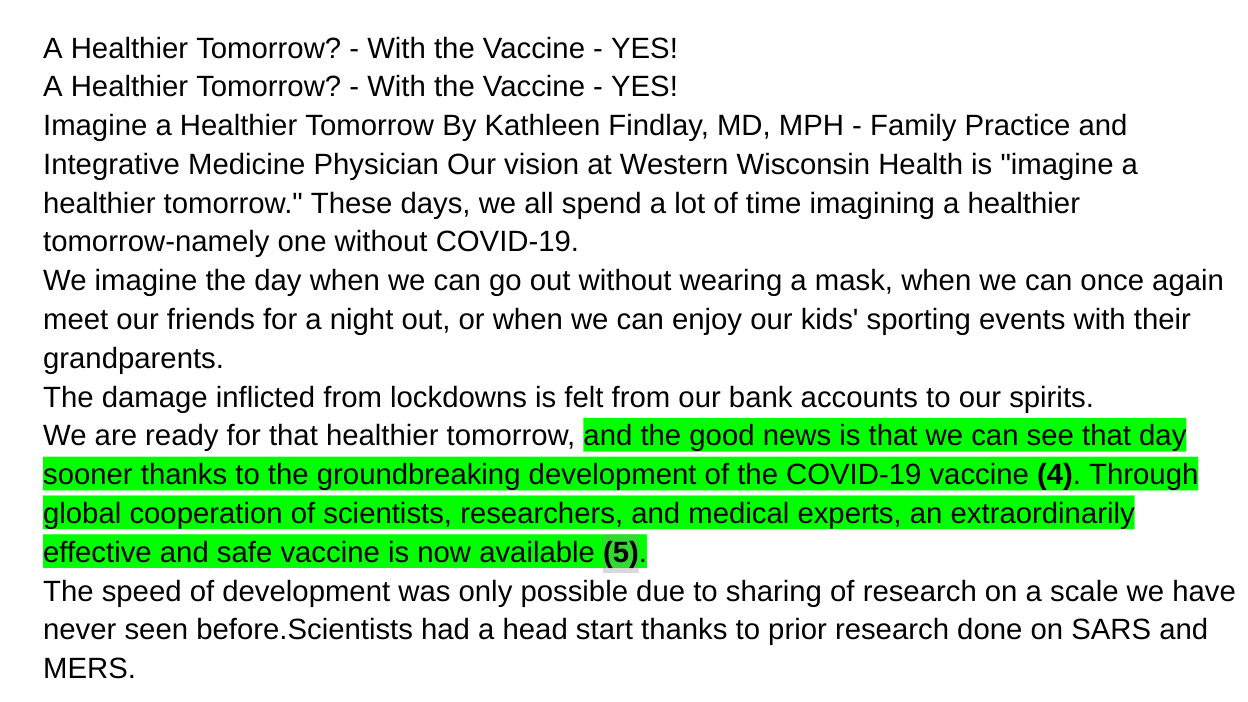}
        \caption{Example 105 from test dataset labeled by our finetuned Longformer with Reasons and their Stance}
    \label{fig:example_A_stance}
\end{figure*}

\begin{figure*}
\centering
        \includegraphics[scale=0.3]{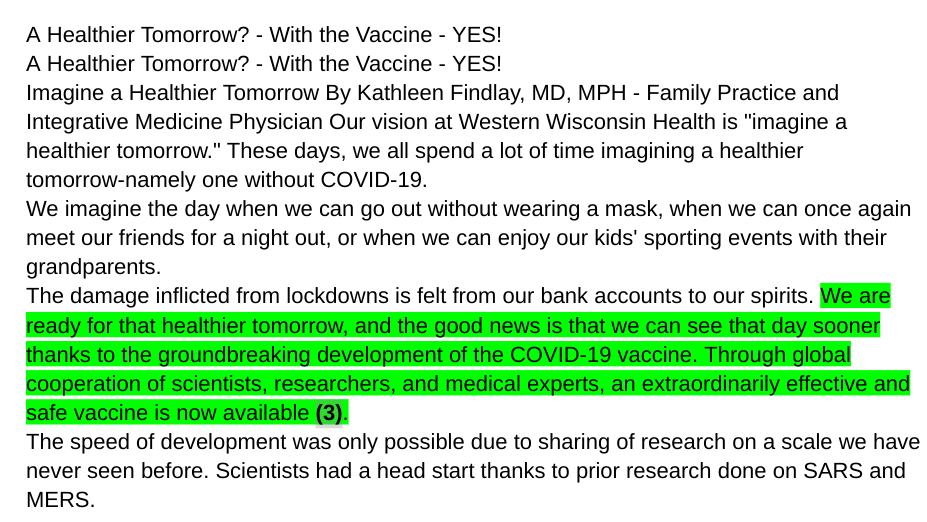}
        \caption{Example 105 from test dataset labeled by our finetuned Longformer with Reasons and the Compressed version of Stances}
    \label{fig:example_A_compressed}
\end{figure*}

\begin{figure*}
\centering
    \centering
    \includegraphics[scale=0.25]{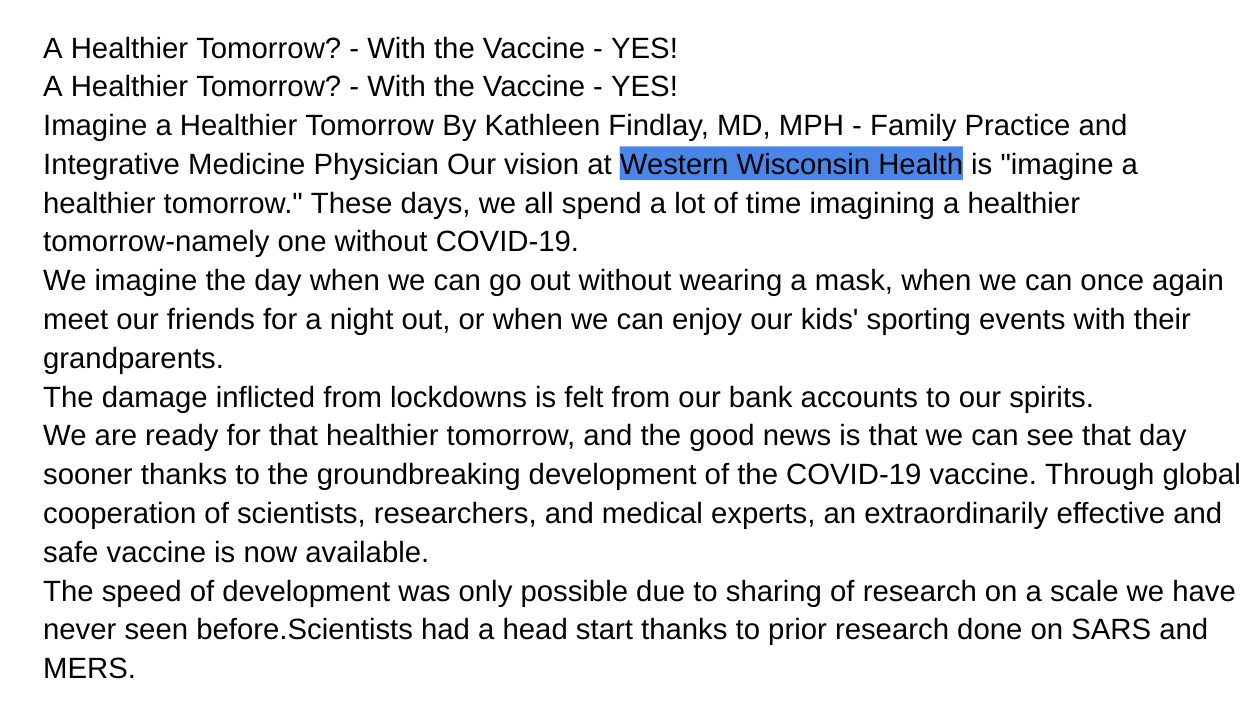}
    \caption{Example 105 from test dataset labeled by our finetuned Roberta-base with Scientific Authorities}
    \label{fig:example_A_scientific_authorities}
\end{figure*}

\begin{figure*}
    \centering
        \includegraphics[scale=0.3]{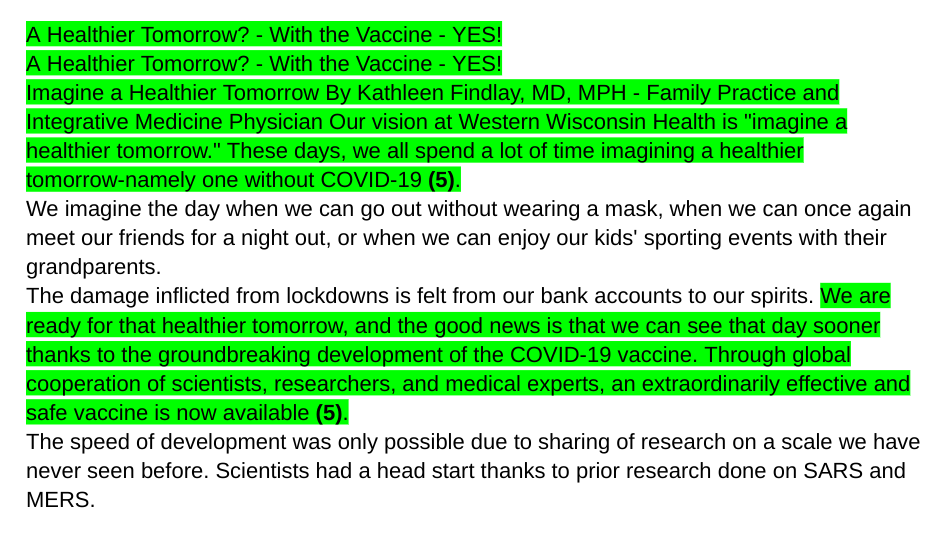}
        \caption{Example 105 from test dataset labeled by GPT4}
    \label{fig:example_A_gpt4}
\end{figure*}

We randomly selected 3 examples from the test set and generated predictions for each of them using GPT4 and the best performing models from each category. Figures \ref{fig:example_A_human}, \ref{fig:example_A_reason}, \ref{fig:example_A_stance}, \ref{fig:example_A_compressed}, \ref{fig:example_A_scientific_authorities} and \ref{fig:example_A_gpt4} show Example 105 from test dataset as annotated by a human annotator, a finetuned Longformer predicting exclusively Reasons, a finetuned Longformer predicting Reasons with their Stance, a finetuned Longformer predicting Reasons with the Compressed Stances, a finetuned Roberta predicting Scientific Authorities, and GPT4 respectively. It can be observed that there is general agreement about the paragraph starting with "We are ready for a healthier tomorrow...", with the exception of the model predicting Stances, which left the first part of it unannotated. Human annotator considered the first two sentences to be valid reasons and GPT4 considered also the other next paragraph, but none of the pretrained models labeled them. The Roberta model predicting Scientific Authorities labeled "Western Winsconsin Health", though Human annotator didn't.

\begin{figure*}
        \centering
        \includegraphics[scale=0.25]{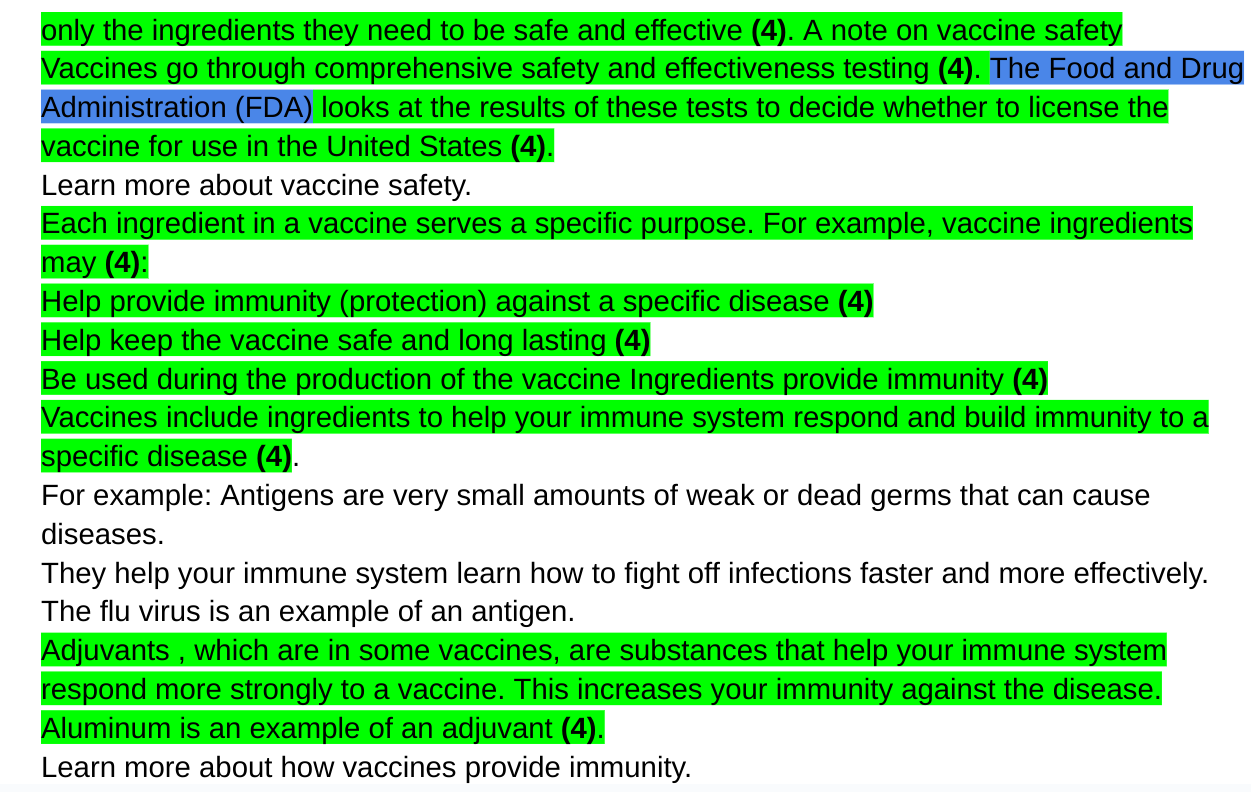}
        \caption{Example 114 from test dataset labeled through nichesourcing with Reasons, Stances and Scientific Authorities}
    \label{fig:example_B_human}
\end{figure*}

\begin{figure*}
        \centering
        \includegraphics[scale=0.25]{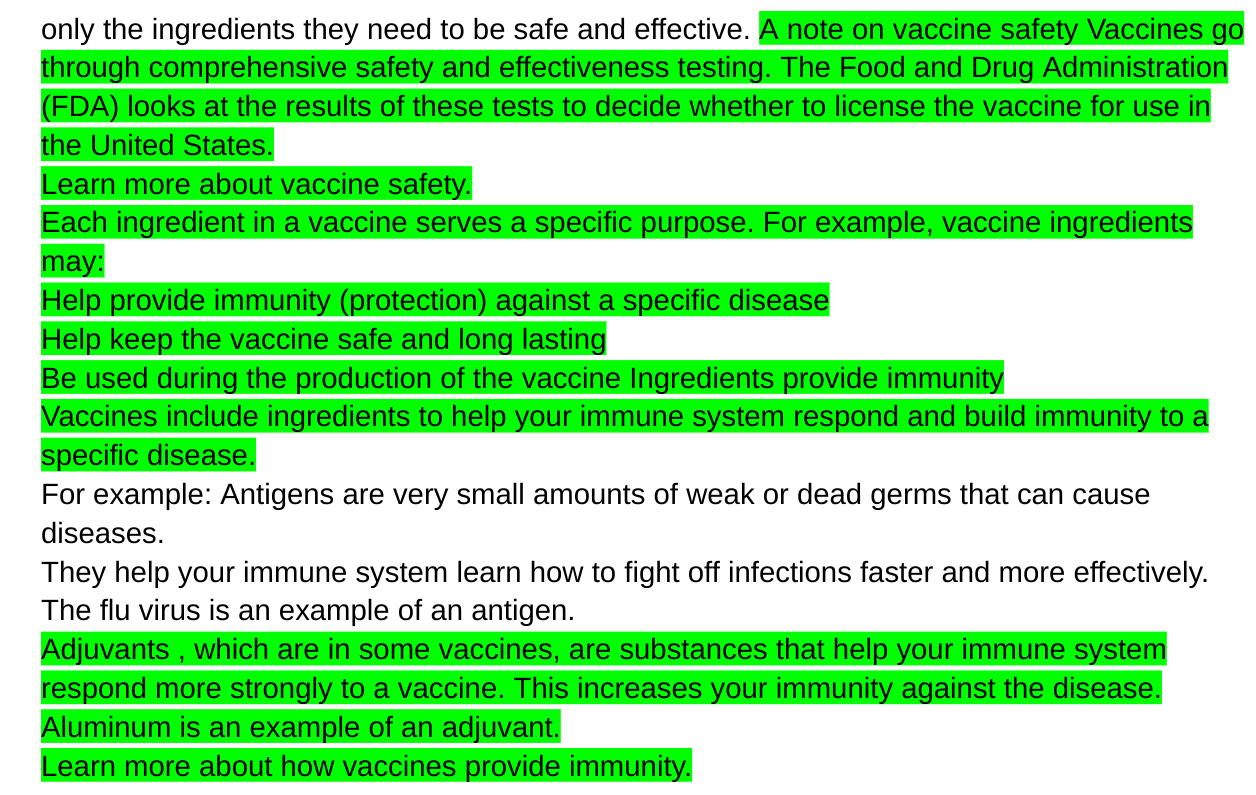}
        \caption{Example 114 from test dataset labeled by our finetuned Longformer for the task of detecting Reasons}
    \label{fig:example_B_reason}
\end{figure*}

\begin{figure*}
        \centering
        \includegraphics[scale=0.25]{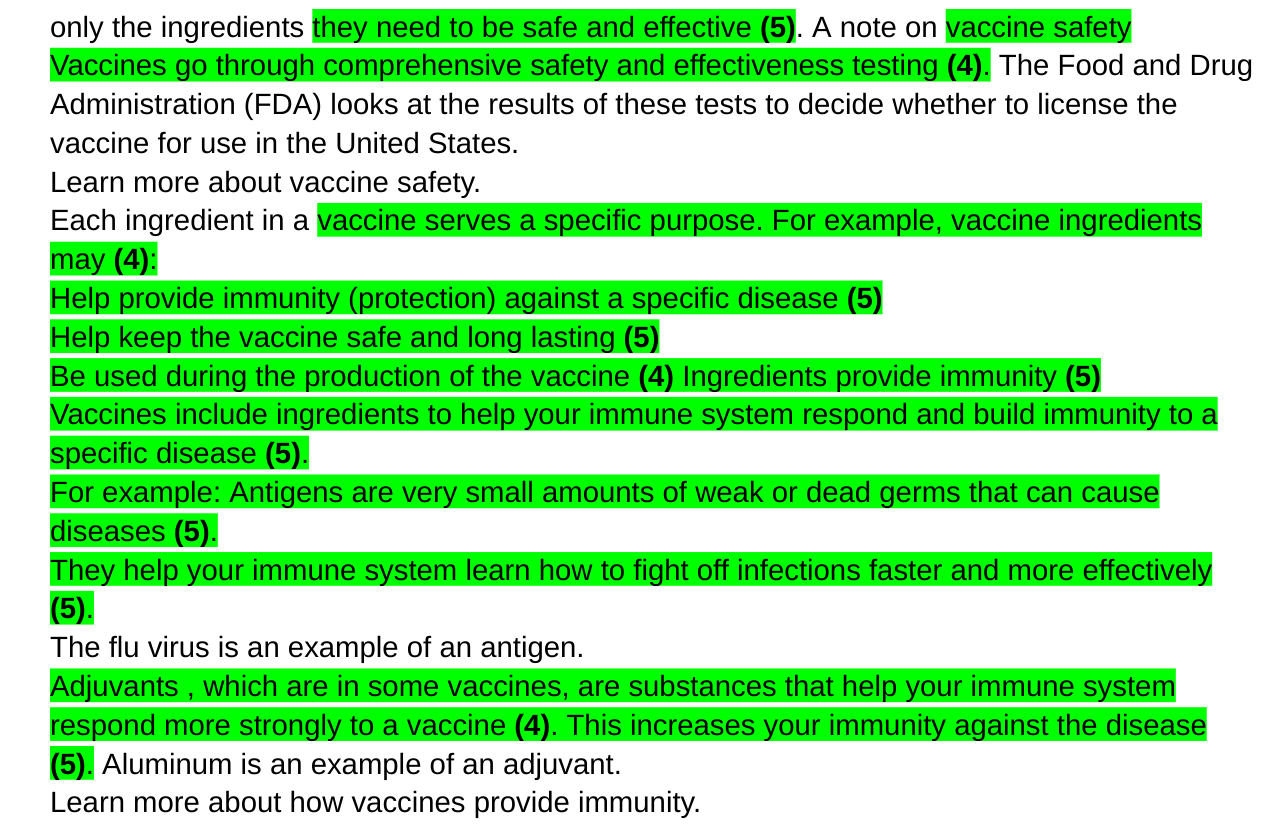}
        \caption{Example 114 from test dataset labeled by our finetuned Longformer for the task of detecting Reasons and their Stance}
    \label{fig:example_B_stance}
\end{figure*}

\begin{figure*}
    \centering
    \includegraphics[scale=0.25]{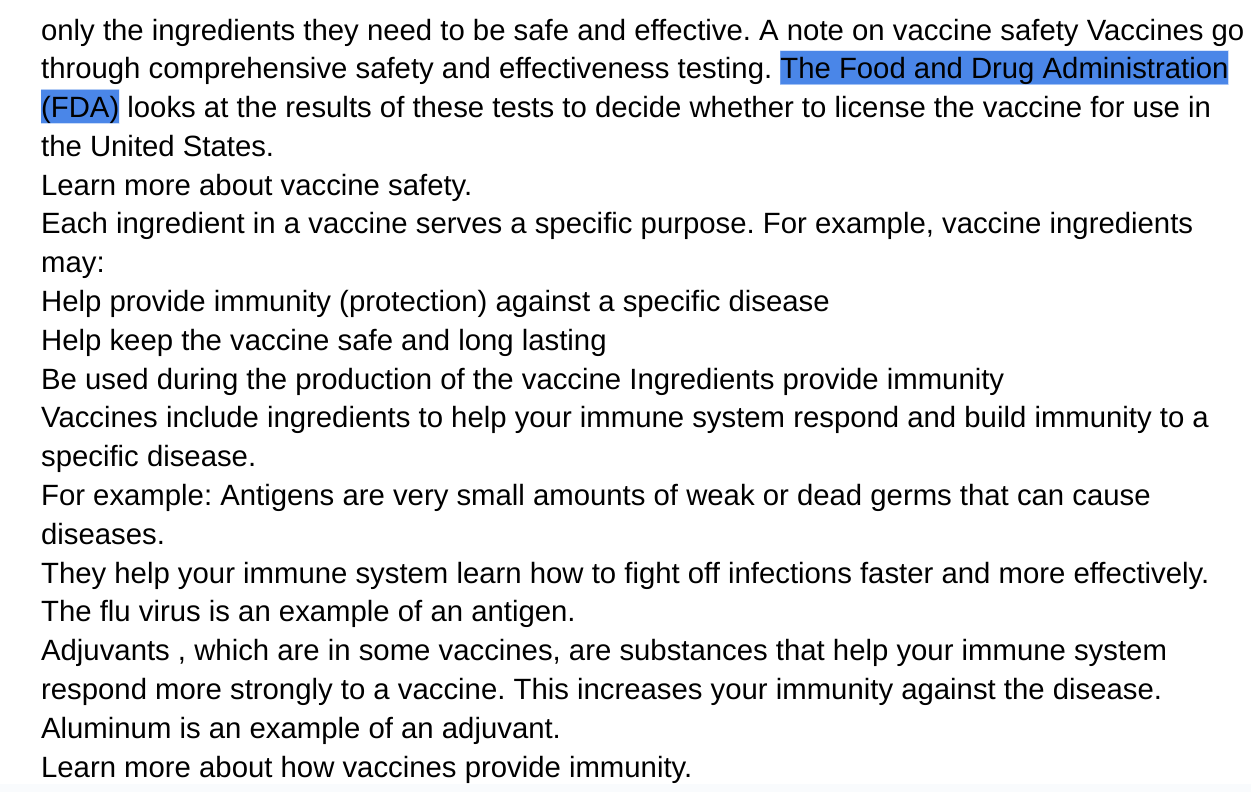}
    \caption{Example 114 from test dataset labeled by our finetuned Roberta-base for the task of detecting Scientific Authorities}
    \label{fig:example_B_scientific_authority}
\end{figure*}

Figures \ref{fig:example_B_human}, \ref{fig:example_B_reason}, \ref{fig:example_B_stance} and \ref{fig:example_B_scientific_authority} show Example 114 from test dataset as annotated, also, by a human annotator, a model predicting exclusively Reasons, a model predicting Reasons with their Stance and a model predicting Scientific Authorities, respectively. In this case, the model predicting Scientific Authorities matched exactly with Human annotation. For the model predicting Reasons, there are minor discrepancies regarding if the phrases starting with "Learn more about..." should be considered a reason or not but mostly matches with Human annotation. When inspecting Stance predictions we observe a higher level of discrepancies regarding what is a reason or not and also about the Stance value though the difference is only between Strong and Weak Support. This example was not labeled by GPT4 because it produced a result that was not correctly labeled (see section \ref{sec:gptresults})

In both examples we found that the Longformer model predicting Stances doesn't label the full extent of a sentence, something that was clearly stated on the annotation protocol and that prevailed on most Human Annotations and also mostly on models predicting only Reasons.

\begin{figure*}
    \centering
    \includegraphics[scale=0.3]{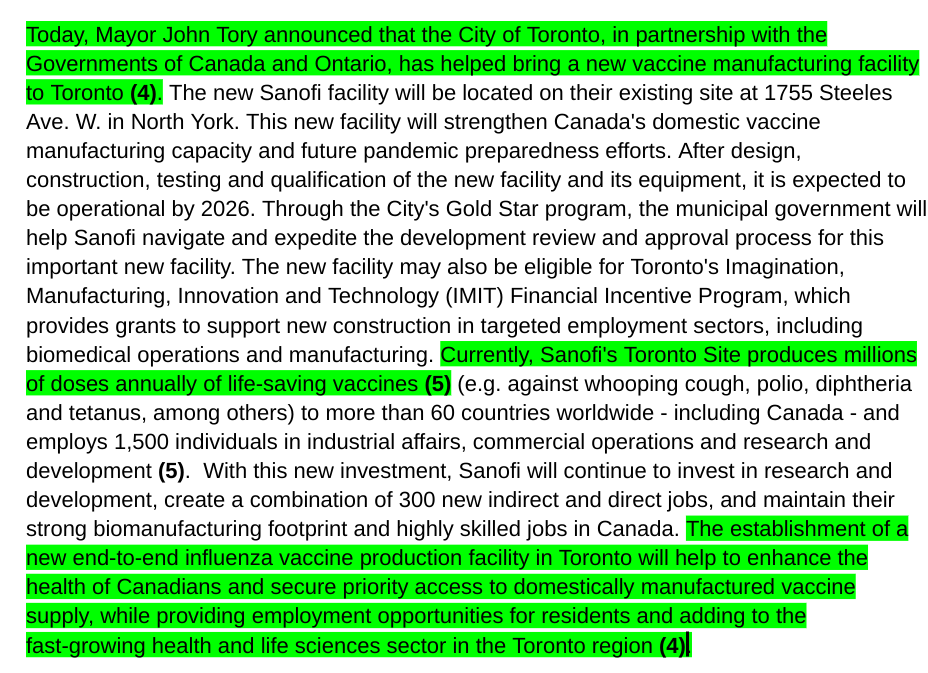}
    \caption{Example 784 from test dataset labeled by a human annotator showing}
    \label{fig:example_784_human}
\end{figure*}

\begin{figure*}
    \centering
    \includegraphics[scale=0.3]{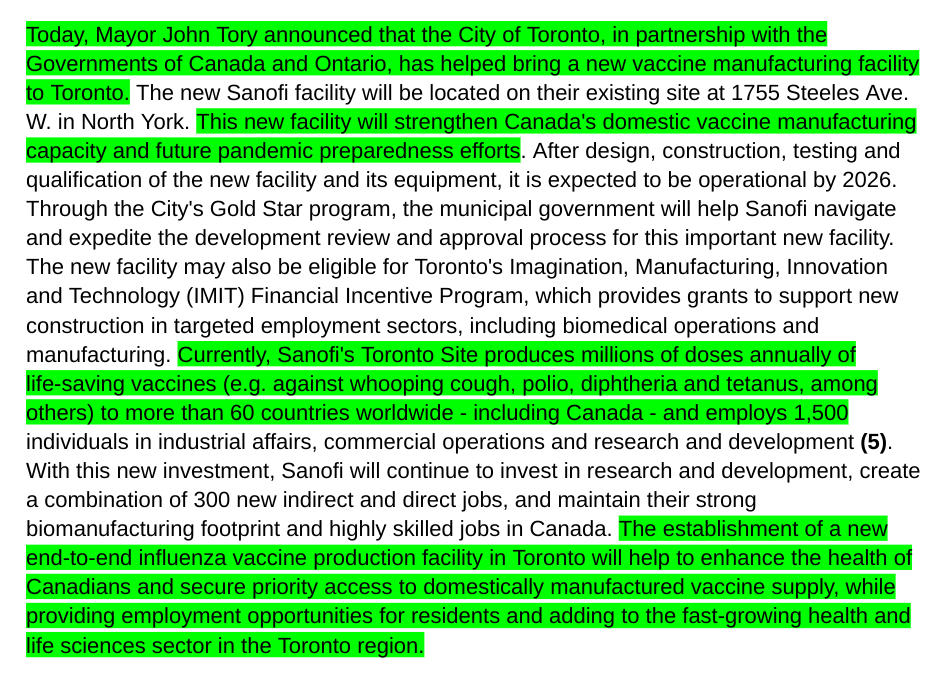}
    \caption{Example 784 from test dataset labeled by a finetuned longformer for the task of detecting Reasons}
    \label{fig:example_784_reasons}
\end{figure*}

\begin{figure*}
    \centering
    \includegraphics[scale=0.3]{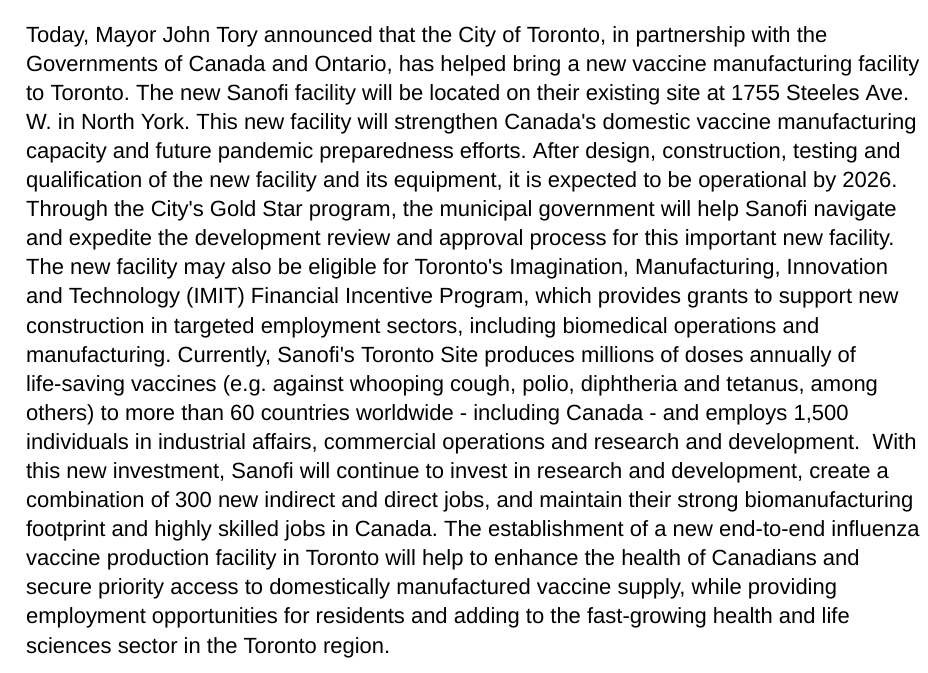}
    \caption{Example 784 from test dataset labeled by a finetuned longformer for the task of detecting Reasons and their Stance}
    \label{fig:example_784_stance}
\end{figure*}

\begin{figure*}
    \centering
    \includegraphics[scale=0.3]{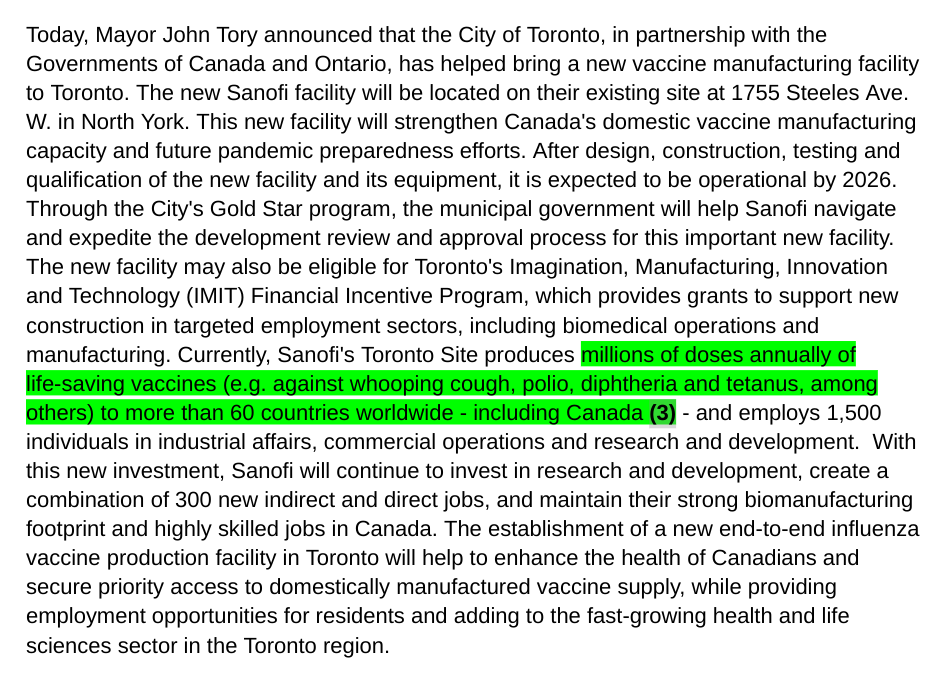}
    \caption{Example 784 from test dataset labeled by a human annotator showing a Reason with a Strong stance Against vaccination}
    \label{fig:example_784_compressed}
\end{figure*}

\begin{figure*}
    \centering
    \includegraphics[scale=0.3]{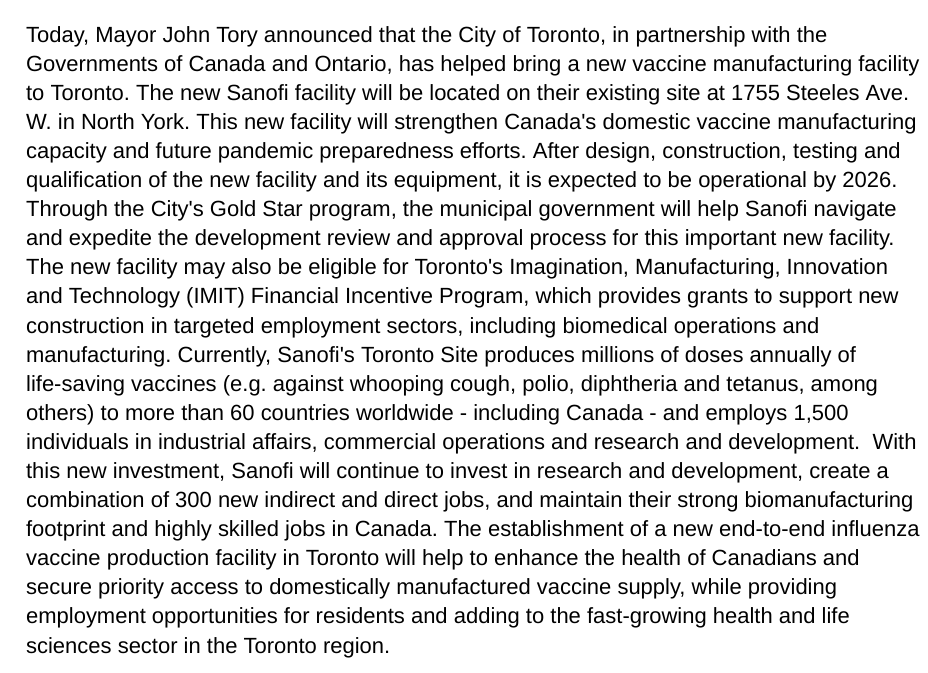}
    \caption{Example 784 from test dataset labeled by a finetuned Roberta model for the task of detecting Scientific Authorities}
    \label{fig:example_784_scientific}
\end{figure*}

\begin{figure*}
    \centering
    \includegraphics[scale=0.3]{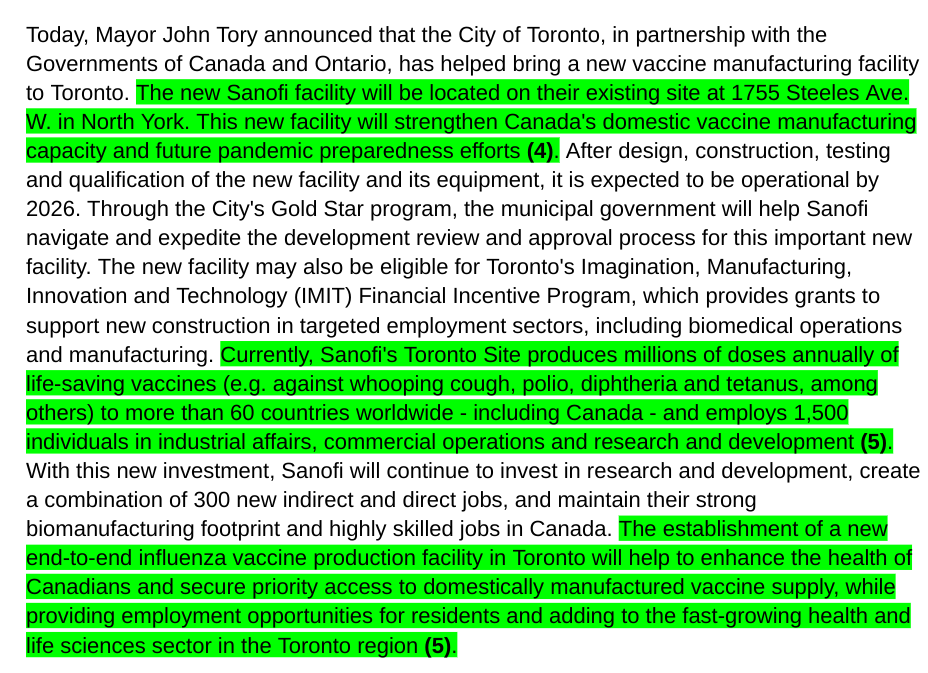}
    \caption{Example 784 from test dataset labeled by a GPT4}
    \label{fig:example_784_gpt4}
\end{figure*}

Figures \ref{fig:example_784_human}, \ref{fig:example_784_reasons}, \ref{fig:example_784_stance}, \ref{fig:example_784_compressed}, \ref{fig:example_784_scientific} and \ref{fig:example_784_gpt4} show Example 784 from test dataset as annotated by a human annotator, a finetuned Longformer predicting exclusively Reasons, a finetuned Longformer predicting Reasons with their Stance, a finetuned Longformer predicting Reasons with the Compressed Stances, a finetuned Roberta predicting Scientific Authorities, and GPT4 respectively. Model predicting only Reasons shows a high level of matching with Human annotations, though it predicted one extra sentence on the first paragraph. The paragraph starting with "Currently, Sanofi's..." was only partially labeled by the Human annotator although the annotation manual clearly states that whenever it was possible, whole sentences should be labeled, just like the Longformer model did. Model predicting Stances didn't make any predictions and thus found no Reasons. However, model predicting the compressed version of the Stances did find a Reason which partially matched one of the Reasons labeled by the Human annotator. No Scientific Authority was labeled on this example neither the model predicted one, so in that sense, they match. Lastly, GPT4 labeled the second sentence and not the first, like the Human annotator did. It labeled the whole paragraph starting with "Currently..." just like the annotation manual says, and also labeled the last paragrapgh matching the annotation made by the Human expert.

\begin{figure*}
    \centering
    \includegraphics[scale=0.25]{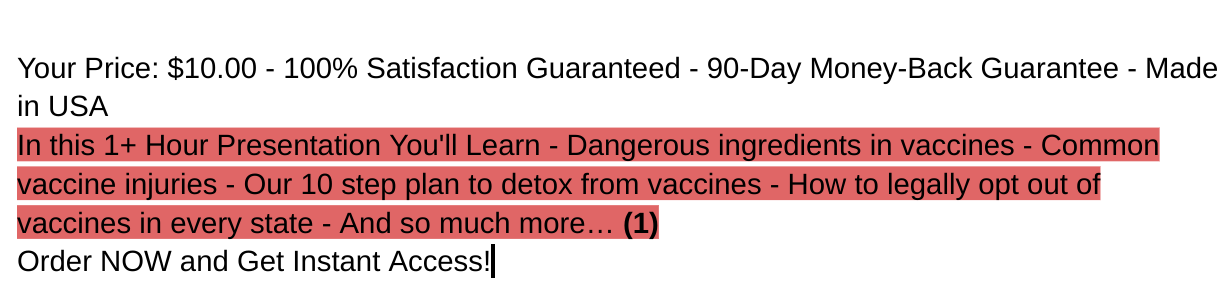}
    \caption{Example 160 from test dataset labeled by a human annotator showing a Reason with a Strong stance Against vaccination}
    \label{fig:example_C}
\end{figure*}

Lastly, figure \ref{fig:example_C} shows example 160 from the test dataset which was manually selected because it has a clear stance against vaccination. In this case, no finetuned model predicted any Reasons or Scientific Authorities. GPT4's prediction, on the other hand, matched exactly with Human annotator.


\begin{figure*}
    \centering
    \includegraphics[scale=0.3]{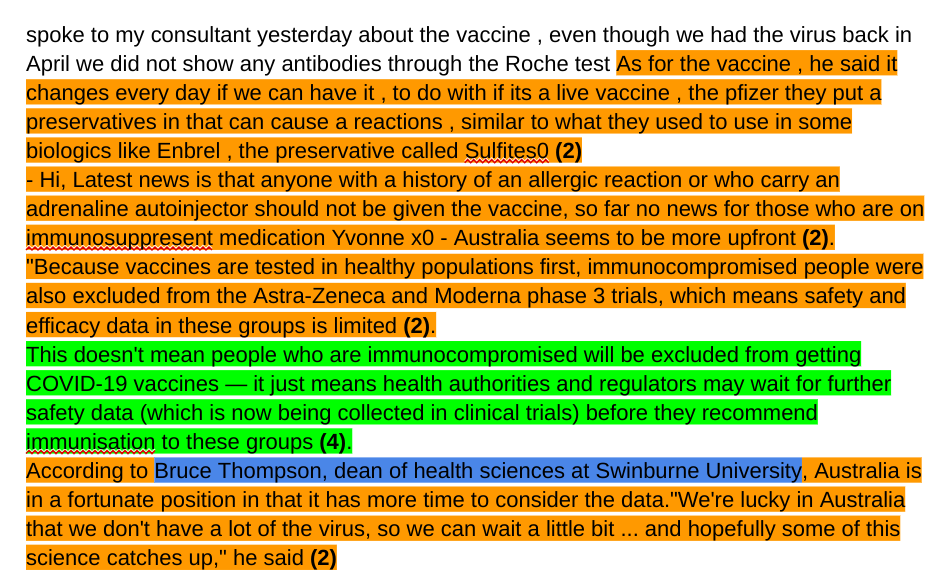}
    \caption{Example annotated by human annotator showing a weak stance against vaccination based on scientific debate about immunosuppressed patients}
    \label{fig:immunosuppressed-patient}
\end{figure*}

\end{document}